\documentclass[a4paper,fleqnw]{cas-sc}
\usepackage{cite}
\usepackage{amsmath,amssymb,amsfonts}
\usepackage{algorithmic}
\usepackage{graphicx,color,subfigure}
\usepackage{textcomp}
\usepackage{booktabs,multirow} 
\usepackage{array} 
\usepackage{paralist} 
\usepackage{enumitem}
\usepackage{verbatim} 
\usepackage[linesnumbered,ruled]{algorithm2e} 
\usepackage{tabularx}  
\usepackage{engord}
\usepackage{url}
\usepackage{comment}
\usepackage{nomencl}

\usepackage{float}
\usepackage{mathtools}
\usepackage{bbding}
\usepackage{amsthm}

\newcommand{\tabincell}[2]{\begin{tabular}{@{}#1@{}}#2\end{tabular}}
\newcommand\cparagraph[1]{\vspace{0.6mm}\noindent\textbf{#1.}}

\usepackage[authoryear,longnamesfirst]{natbib}

\def\tsc#1{\csdef{#1}{\textsc{\lowercase{#1}}\xspace}}
\tsc{WGM}
\tsc{QE}
\tsc{EP}
\tsc{PMS}
\tsc{BEC}
\tsc{DE}

\begin{document}
\let\WriteBookmarks\relax
\def\floatpagepagefraction{1}
\def\textpagefraction{.001}

\shorttitle{Strategic Coordination of Drones via Short-term Distributed Optimization and Long-term Reinforcement Learning}

\shortauthors{Chuhao Qin et~al.}

\title [mode = title]{Strategic Coordination of Drones via Short-term Distributed Optimization and Long-term Reinforcement Learning}                      

\author[1]{Chuhao Qin}[orcid=0000-0002-6178-7973]
\cormark[1]
\ead{sccq@leeds.ac.uk}

\author[1]
{Evangelos Pournaras}
\ead{e.pournaras@leeds.ac.uk}

\affiliation[1]{organization={School of Computer Science, University of Leeds},
    city={Leeds},
    country={UK}}

\cortext[cor1]{Corresponding author: Chuhao Qin (email: sccq@leeds.ac.uk).}

\begin{abstract}
This paper addresses the problem of autonomous task allocation by a swarm of autonomous, interactive drones in large-scale, dynamic spatio-temporal environments. When each drone independently determines navigation, sensing, and recharging options to choose from such that system-wide sensing requirements are met, the collective decision-making becomes an NP-hard decentralized combinatorial optimization problem. Existing solutions face significant limitations: distributed optimization methods such as collective learning often lack long-term adaptability, while centralized deep reinforcement learning (DRL) suffers from high computational complexity, scalability and privacy concerns. To overcome these challenges, we propose a novel hybrid optimization approach that combines long-term DRL with short-term collective learning. In this approach, each drone uses DRL methods to proactively determine high-level strategies, such as flight direction and recharging behavior, while leveraging collective learning to coordinate short-term sensing and navigation tasks with other drones in a decentralized manner. Extensive experiments using datasets derived from realistic urban mobility demonstrate that the proposed solution outperforms standalone state-of-the-art collective learning and DRL approaches by $27.83\%$ and $23.17\%$ respectively. Our findings highlight the complementary strengths of short-term and long-term decision-making, enabling energy-efficient, accurate, and sustainable traffic monitoring through swarms of drones.
\end{abstract}



\maketitle

\section{Introduction}
Unmanned Aerial Vehicles (UAVs), referred to as drones, can organize themselves into swarms, fostering collaboration and efficiency in sensor data collection within Smart Cities. With their mobility, autonomy, and intelligence, drones have been widely used in several applications, such as monitoring traffic congestion~\cite{barmpounakis2020new}, mapping natural disasters~\cite{deng2020two} and smart agriculture~\cite{liu2024complex}. Despite these advantages, drones are constrained by limited battery capacity, which impacts their spatio-temporal coverage. This limitation necessitates sophisticated autonomous control systems and precise task allocation within drone swarms. This involves the allocation of different sensing tasks to each drone while meeting the sensing requirements, drone capabilities and constraints~\cite{poudel2022task}. Recently, scholars have solved the NP-hard combinatorial optimization problem of assigned sensing tasks by drone swarms, using task allocation algorithms ranging from genetic algorithms to swarm intelligence methods~\cite{poudel2022task,zhou2020uav}. 

To enhance robustness and minimize the effects of individual drone failures, distributed optimization approaches, such as decentralized collective learning, have been introduced~\cite{pournaras2018decentralized}. These approaches coordinate agents to autonomously self-organize and self-assign task plans in a decentralized manner, respecting the autonomy and operational flexibility of individual agents. Due to their scalability (supporting a large number of agents), efficiency (low communication and computational cost), decentralization and resilience~\cite{pournaras2018decentralized}, they are well-suited for tackling the autonomous task allocation problem for large-scale spatio-temporal sensing (i.e., a decentralized combinatorial optimization problem), which is the focus of this paper. In particular, they solve the navigation (determining flight path), sensing (scheduling data collection) and recharging (choosing charging stations as destinations) of drone swarms. This approach is especially valuable in scenarios involving multi-modal sensing requirements, where heterogeneous drones, equipped with diverse sensors (e.g., camera, LiDAR, thermal), collaborate across different institutions to perform coordinated sensing while preserving the privacy of individual agents.

However, distributed optimization approaches are reactive to the current state of the environment rather than proactive to address strategically long-term operational challenges. When the time horizon for autonomous task allocation by drone swarms extends-from short durations constrained by flight range to longer periods spanning a whole day-the sensing environment (e.g., traffic flow) becomes increasingly dynamic and unpredictable. To remain effective in such evolving conditions, drone swarms must incorporate strategic foresight into their decision-making, i.e., anticipating flying directions and future sensor data collective requirements. For example, drones aware of an expected increase in traffic can proactively fly to those areas, even if those areas are currently sparsely populated with vehicles. They can also land on base stations near those areas for recharging and executing sensing tasks in the next time period. As a result, the strategic decision-making improves tasks of vehicle detection, even if the short-term benefits are not evident. This "slower is faster" effect emphasizes the importance of long-term planning for overall effectiveness in drone sensing. 

To address long-term optimization, deep reinforcement learning (DRL) has been introduced, utilizing the Bellman equation to consider cumulative rewards~\cite{cui2019multi}. DRL based on deep neural networks support drones, as heterogeneous agents, to manage complex state spaces and adapt to changing environments~\cite{cui2019multi}. Nevertheless, traditional DRL approaches struggle with the autonomous task allocation problem because of their inherent scalability limitations, high complexity, and reliance on centralized control. As sensing tasks prolong to an entire day, the number of decisions per drone increases significantly since DRL typically operates on short time intervals (e.g., seconds or minutes). This leads to an exponential growth of the action space, increasing training complexity and hindering effective exploration of optimal action to allocate sensing tasks. The curse of dimensionality could be furt her exacerbated by a larger number of agents/drones. Moreover, centralized training requires access to private or sensitive information of drones, including battery level and collected sensor data, thereby compromising the autonomy and privacy of heterogeneous drones. Notably, these limitations of DRL are where distributed optimization methods excel, offering scalable, decentralized, and privacy-preserving solutions, motivating the study of a hybrid approach.

Therefore, in this paper, we propose a new hybrid optimization approach to coordinate drone swarms through Distributed Optimization and (deep) Reinforcement Learning (\emph{DO-RL}), which involves a multi-agent DRL algorithm that builds upon on a decentralized collective learning approach~\cite{pournaras2018decentralized}. This approach leverages a hierarchical decision-making framework that uses both high-level long-term policies and low-level short-term coordination. Specifically, \emph{DO-RL} allows drones to determine flying directions from departures to destinations (i.e., charging stations) over multiple time periods via DRL, while autonomously selecting their navigation and sensing plans within a time period through collective learning. The plans are generated locally and autonomously by drones without leaking their private data to others. As a consequence, the low-level collective learning not only reduces the number of actions in high-level DRL and training complexity, but efficiently coordinates each drone to select plans and observe the aggregated plans of others through a tree communication topology. The aggregated plans are used to calculate the reward of DRL for optimizing sensing performance.

The contributions of this paper are outlined as follows: 
(i) The first study of autonomous task allocation for large-scale, dynamic spatio-temporal sensing by a swarm of drones, concentrating on optimizing the flight paths, data collection strategies, and recharging. (ii) A novel hybrid optimization approach, \emph{DO-RL}, that tackles this problem by integrating both multi-agent DRL and decentralized collective learning. It includes a set of auxiliary methods such as DRL-based scheduling, local plan generation, iterative plan selection, periodic state update, actor-critic networks, and policy proximal optimization. (iii) A testbed for extensive experimentation with realistic traffic patterns and real-world transport networks to validate the higher energy-efficient and accurate vehicle observation of \emph{DO-RL} compared to state-of-the-art baseline methods. (iv) Quantitative findings that show how the proposed hybrid approach achieves a win-win synthesis of long-term learning and short-term optimization in both sensing and charging under a large spectrum of factors such as drone density, number of time periods, and vehicle density. (v) An open-source implementation of the proposed approach and an open dataset\footnote{Available at: https://doi.org/10.6084/m9.figshare.24476533.v2} containing all plans of the studied scenario. They can be used as benchmarks to encourage further research on this problem.


\section{Related Work} \label{sec:related}
The sensing task allocation problem of drones has been traditionally defined as the traveling salesmen problem or vehicle routing problem, which is a NP-hard combinatorial optimization problem~\cite{bartolini2019task}. This involves finding the optimal assignment of tasks to drones at specific locations, while considering constraints such as task urgency, time scheduling, and flying costs. The task allocation algorithms, such as Particle Swarm Optimization (PSO)~\cite{phung2021safety}, and genetic algorithm~\cite{pehlivanoglu2021enhanced}, have been introduced to formulate the problem of reaching optimal sensing efficiency in terms of coverage, events detection rate and the cost of flight paths. They rely on adaptive exploration of the solution space to optimize task allocation based on shared information and local interactions. However, these methods are centralized, wherein the failure of a central control station can lead to several system disruptions without autonomous recovery mechanisms~\cite{liu2023decentralized}. 

To solve the decentralized combinatorial optimization problem, several distributed task allocation algorithms have been proposed: One such algorithm is the Robust Decentralized task allocation (RDTA)~\cite{alighanbari2006robust} that enhances robustness and reduces communication costs by employing decentralized planning for drones.  Another one is the Consensus-Based Bundle Algorithm (CBBA)~\cite{bertuccelli2009real}, which combines market-based mechanisms and situation awareness to converge and avoid task conflicts among drones. The COHDA~\cite{hinrichs2013cohda} optimization generalizes well in different communication structures among drones with full view of the system, but requires each drone to exchange information with all of others. Moreover, a decentralized multi-agent collective learning approach, named Economic Planning and Optimized Selections (EPOS), allows agents to exchange information over a tree-structured communication topology, achieving low computational and communication overhead~\cite{pournaras2018decentralized}. Compared to the relevant algorithm of COHDA, this approach has lower cost of message exchange and higher cost-effectiveness in terms of optimality~\cite{pournaras2018decentralized}. Moreover, EPOS enables each drone to share the aggregated plans of navigation and sensing to others, without exposing the private information of drone specifications, battery and sensor data~\cite{qin2024m}. Nevertheless, long-term sensing scenarios pose challenges. Drones tend to prioritize immediate maximum returns in the short term, often neglecting the need for strategic decision-making for long-term benefits.

The deep reinforcement learning (DRL) algorithm proves to be effective for addressing  the combinatorial optimization problem in the evolving multi-drone sensing environments, by transforming the problem in to a sequential decision-making process~\cite{barrett2020exploratory,mazyavkina2021reinforcement}. Ding \textit{et al.}~\cite{ding2021crowdsourcing} integrate mobile crowdsensing (i.e., human equipped with mobile sensing devices) into UAV sensing, overcoming the limitations of both human mobility and drone battery life. They use the centralized training decentralized execution framework where drones can share auxiliary information during training but act independently during execution, preserving decentralization. Zhao \textit{et al.}~\cite{zhao2022dronesense} propose a comprehensive solution for sustainable urban-scale sensing in open parking spaces. Their solution incorporates the task selection and scheduling using DRL, and adaptive recharging assignment. Omoniwa \textit{et al.}~\cite{omoniwa2023communication} present a multi-agent decentralized DRL approach that improves drone energy efficiency in providing wireless connectivity by sharing information with neighboring drones. However, these approaches are confined to a small-scale task allocation problem due to high training complexity.

The scalability of multi-agent DRL has been widely studied. Previous work combines DRL with mean-field methods~\cite{yang2018mean,chen2020mean} that transforms the complex interactions within the population of agents into a simplified interaction between a single agent and the average effect from the overall population. Other work focus on reducing the joint action-state space to improve the exploration efficiency~\cite{kanervisto2020action,majeed2021exact,machado2023temporal}. Apart from the action abstraction approaches that rely on predefined and offline abstractions, such as masking, sequentialization and temporal abstraction, the hierarchical reinforcement learning~\cite{jendoubi2023multi,xu2023haven} has been introduced to provide flexibility in handling non-linear and complex task environments. It structures decision-making into high- and low-level tasks, enabling agents to learn complex behaviors more efficiently by decomposing the problem. Furthermore, a recent work by Qiu \textit{et al.}~\cite{qiu2022dimes} propose DIMES, a differentiable meta-solver that leverages the compact continuous space and DRL to address traveling salesmen problems at large scales (up to 10000 nodes in a graph). Nevertheless, these approaches lack autonomy for drones in autonomous allocation tasks, increasing vulnerability to agent failures and private information leakage.

In the application of transportation system, drones rely on accurate vehicle observation to detect early signs of traffic congestion. This enables traffic operators to apply mitigation measures, reducing the carbon footprint in one of the world's highest-emission sectors. Samir \textit{et al.}~\cite{samir2020age} leverage a DRL algorithm to optimize the trajectories of UAVs, with the objective of minimizing the age of information of collected data. Bakirci \textit{et al}.~\cite{bakirci2024enhancing} explore vehicle detection using the YOLOv8 algorithm on aerial images captured by a custom UAV. While these approaches demonstrate high performance in the detection of traffic vehicles, scalability becomes a challenge since they only deal with a small number of drones within a short duration. 

\begin{table}[!t]
	\centering
	\caption{ Comparison to related work: criteria covered (\Checkmark) or not (\XSolid).}  
	\label{table:criteria}
    \resizebox{13cm}{!}
    {
     	\begin{tabular}{lcccccccc}  
		\toprule  
		\textbf{Criteria \, Ref.:} &\tabincell{l}{\cite{bertuccelli2009real}} &\tabincell{l}{\cite{hinrichs2013cohda}} &\tabincell{l}{\cite{qin2024m}} &\tabincell{l}{\cite{ding2021crowdsourcing}} &\tabincell{l}{\cite{zhao2022dronesense}} &\tabincell{l} 
        {\cite{qiu2022dimes}} &\tabincell{l} {\cite{samir2020age}} &This paper\\  
		\midrule     
    	Decentralization		&\Checkmark    &\Checkmark	&\Checkmark	&\Checkmark	&\XSolid	   &\XSolid      &\XSolid	&\Checkmark \\
  
  	Long-term adaptability 	&\XSolid   &\XSolid &\XSolid &\Checkmark &\Checkmark &\Checkmark &\Checkmark &\Checkmark \\
    
		Scalability with low complexity		&\Checkmark &\Checkmark	&\Checkmark	&\XSolid	&\XSolid	&\Checkmark &\XSolid	&\Checkmark \\  

        Autonomy and privacy-preservation     &\XSolid &\XSolid	&\Checkmark	&\XSolid	&\XSolid	&\XSolid &\XSolid	&\Checkmark \\ 


  
		\bottomrule
	\end{tabular}  
    }

\end{table} 

In summary, previous approaches have either neglected the long-term advantages of strategic sensing, or failed to scale without encountering centralized bottlenecks. To overcome these shortcomings, the proposed approach empowers drones to make informed and proactive decisions based on predicted traffic flow, while enabling them to independently adjust to changing conditions and unexpected events, ensuring both scalability and adaptability without relying on centralized control. Table~\ref{table:criteria} illustrates the comparison to related work in different types of criteria.

\section{System Model} \label{sec:model}
In this section, we define key concepts of scenarios and then formulate the main performance metrics. Table~\ref{table:notation} illustrates the list of mathematical notations used in this paper.

\begin{table}[!t]
	\caption{Notations.}  
	\centering
	\begin{tabularx}{\linewidth}{lXl}  
		\hline  
		Notation & Explanation \\  
		\hline     
		$ u, U, \mathcal{U} $  & Index of a drone; total number of drones; set of drones \\  
    	$ m, M, \mathcal{M} $  & Index of a charging station; total number of charging stations; set of charging stations \\  
		$ n, N, \mathcal{N} $  & Index of a grid cell; total number of grid cells; set of grid cells \\
		$ s, S, \mathcal{S} $  & Index of a timeslot; total number of timeslots in a period; set of timeslots \\ 
        $ t, T, \mathcal{T} $  & Index of a period; total number of periods; set of periods \\ 
        $ a^u $  & The action or flying direction taken by $u$ \\
        $ P^u, p^u_{n,s} $  & The plan of $u$; value of $P^u$ at cell $n$ and timeslot $s$ \\
        $ P^{-u}, p^{-u}_{n,s} $  & The observed plan of drones by $u$ excluding $P^u$; value of $ P^{-u}$ at cell $n$ and timeslot $s$ \\
        $ p_{n,s}$  & The aggregated plans of all drones at cell $n$ and timeslot $s$\\
        $ V_{n,s} $  & Required sensing value at cell $n$ and timeslot $s$ \\ 
        $ v_{n,s} $  & Collected sensing value at cell $n$ and timeslot $s$ \\ 
        $ \mathcal{R}, r_{n,s} $  & Target; target value at cell $n$ and timeslot $s$  \\ 
        $ l, L $  & Index of a plan; total number of plans \\ 
        $ C^\mathsf{f}(u), t^\mathsf{f} $ & Flying power consumption of $u$; flying time \\
        $ C^\mathsf{h}(u), t^\mathsf{h}_u $ & Hovering power consumption of $u$; hovering time \\
        $ B(u) $ & Battery capacity of $u$\\
        $ \alpha_1, \alpha_2, \alpha_3 $ & Weight parameters of mission efficiency, sensing accuracy and energy cost in the main objective function respectively\\ 
        $ c^u, e^u $  & Current location of $u$; current energy consumption of $u$\\
        $ K(a^u) $ & Indexes of cells within the searching range along flying direction $a^u$\\ 
        $ J(a^u) $  & Indexes of visited cells within $K(a^u)$\\
        $ \beta $  & Behavior of an agent in planning optimization \\
        $ o^u_t, r^u_t $  & Observation of $u$ at period $t$; reward of $u$ at the period $t$ \\
        $ k, H $ & Index of a sampled transition; number of sampled transitions\\
        $ \gamma, \epsilon $ & Discount factor; clip interval hyperparameter \\
        $ Q(\cdot), \theta^Q $  & Critic network; parameter of critic network \\
        $ \pi(\cdot), \theta^\pi $  & Actor network; parameter of actor network \\
        $ \omega_{n,s}, \overline{v} $ & Prediction coefficient to calculate $\hat{V}_{n,s}$; threshold to remove the regions with low sensing requirements \\
        $ I, \mathcal{X}, A $  & Total number of iterations in \emph{EPOS}; size of state space; number of actions in DRL \\
        $ \mathcal{E}, W $  & Number of episodes; number of neutrals per layer in DNN \\
		\hline  
	\end{tabularx}  
	\label{table:notation}
\end{table} 


\subsection{Scenario and assumptions}
Consider a swarm of drones $\mathcal{U} \triangleq \{1,2,...,U\}$ performing sensing missions, such as monitoring vehicles, over a grid that represents a 2D map. In this scenario, a set of grid cells (or points of interest) $\mathcal{N} \triangleq \{1,2,...,N\}$ are uniformly arranged to cover the map. The primary goal of the drones is to coordinate their visits to these cells to collect the required data. Furthermore, a set of charging stations $\mathcal{M} \triangleq \{1,2,...,M\}$, from which the drones depart from and return to, are located at fixed coordinates on the map. In addition, the whole time span is divided into a set of time periods as $\mathcal{T} \triangleq \{1,2,...,T\}$. Each time period can be divided into a set of equal-length scheduling timeslots $\mathcal{S} \triangleq \{1,2,...,S\}$. In each timeslot, a drone can be controlled to fly to a cell and hover to collect sensor data. To explain the sensing performance of drones within each time period, several definitions are illustrated, including action, plans, required sensing values and target.

\cparagraph{Action} To explain the long-term navigation and sensing over all periods, the period-by-period actions of each drone $a^u = \{0, 1, 2,..., 8\}$ is introduced, $u \in \mathcal{U}$, meaning to control $u$ to move horizontally along eight directions, which are 1 = north (N), 2 = east (E), 3 = south (S), 4 = west (W), 5 = northeast (NE), 6 = southeast (SE), 7 = southwest (SW), and 8 = northwest (NW), or return to the origin ($a^u = 0$). Under each action, $u$ executes a short-term navigation and sensing. After completing its sensing tasks, $u$ flies back to one of the charging stations to recharge fully and resume work in the next period $t + 1$.

\cparagraph{Matrix of plans} To explain the short-term navigation and sensing of drones over the cells and timeslots in a period $t$, $t \in \mathcal{T}$, the plan of a drone $u$ that travels through the direction of $a^u$ is introduced, denoted by $P^u(a^u, t)$. It represents the specific navigation and sensing details, including the visited cells and corresponding energy consumption. The plan $P^u(a^u, t)$ is encoded by a matrix of size $N \times S$, with each element represented as $p^u_{n,s}(a^u, t) \in \{0, 1\}$. Here, $p^u_{n,s}(a^u, t) = 1$ denotes that the drone $u$ hovers and collects all required data at cell $n$ at timeslot $s$, whereas for $p^u_{n,s}(a^u, t) = 0$ the drone does not hover at that cell at that time. 
Moreover, $P^{-u}(a^u, t) = \{p^{-u}_{n,s}(a^u, t) |_{n \in \mathcal{N}, s \in \mathcal{S}} \}$ denotes the observed plan by drone $u$ in time period $t$, indicating that it incorporates and sums the plans of all other drones excluding its own. The aggregated plans of all drones observed by $u$ at cell $n$ and timeslot $s$ is formulated as follows:
\begin{equation}
    p_{n,s}(a^u, t) = p^u_{n,s}(a^u, t) + p^{-u}_{n,s}(a^u, t).
    \label{eq:aggregated_plan}
\end{equation}

\cparagraph{Matrix of required sensing values} In the context of a sensing task, each cell at a timeslot has specific sensing requirements that determine the data acquisition goal of drones. Such sensing requirements can be determined by city authorities as a continuous kernel density estimation. The high risk level of a cell at a timeslot represents the high importance of sensing (e.g., the crucial intersection of traffic flow), and thus a high number of required sensing values is set. The matrix of required sensing values is denoted as $V(t) = \{V_{n,s}(t) |_{n \in \mathcal{N}, s \in \mathcal{S}} \}$, where $V_{n,s}(t)$ denotes the required sensing value at cell $n$ and timeslot $s$ in the time period $t$. Based on actual sensing performance of drones according to Eq.(\ref{eq:aggregated_plan}), the sensing value collected by all drones observed by $u$ at cell $n$ and timeslot $s$ is formulated as:
\begin{equation}
    v_{n,s}(a^u, t) = p_{n,s}(a^u, t) \cdot V_{n,s}(t).
\end{equation}

\cparagraph{Matrix of target} In a real-world scenario, drones lack prior knowledge of the amount of required sensing values before they begin their sensing operations. Therefore, it becomes essential to build a target on the fly that instructs the drones regarding when and where they should or should not fly to given information from the environment. 
The matrix of the target is defined as $\mathcal{R}(t)$, which denotes the sensing requirements for all drones at period $t$. The element of the target is denoted as $r_{n,s}(t) = \{0,1\}$, $n \in \mathcal{N}$, $s \in \mathcal{S}$, where $r_{n,s}(t) = 1$ requires only a drone to visit the cell $n$ at timeslot $s$, and $r_{n,s}(t) = 0$ does not require sensor data collection by a drone. Therefore, the system needs to find a optimal combination of the aggregated plans to meet the target. The number of combinations is $O(L^U)$, where $L$ denotes the total number of plans. The overall problem can be classified as NP-hard $0-1$ multiple-choice combinatorial optimization problem, which is decentralized when each drone self-determines its own plan.

Note that our model addresses task allocation problem for drone swarms, i.e., determining \textit{what tasks to perform} and \textit{where to sense}, rather than focusing solely on control and communication strategies that govern \textit{how tasks are executed} (e.g., collision avoidance or minimizing latency). Therefore, to simplify the scenario, this paper makes the following assumptions: 
(i) Each drone is programmed to fly at a distinct altitude during its movement between cells to prevent mid-air collisions risks~\cite{tang2021systematic}. While previous work~\cite{qin2024m,batinovic2022path} has explored a safe and cost-effective approach to a more realistic collision avoidance, this is not the primary focus of this paper.
(ii) Each time period concludes a flying period and a charging period: the time that drones perform sensing, and the time for charging. All drones finish charging before the next time period begins. We assume that each charging period is of equal duration and provides sufficient time for the drones to be fully charged.
(iii) Each charging station is adequately equipped with charging capacity, enabling multiple drones to charge simultaneously without the need for queuing. This arrangement prevents any delays in the charging process.
(iv) Drone computations are offloaded to a remote edge–cloud infrastructure, where decisions are made based on data reliably transmitted from the drones via stable communication links~\cite{nezami2025computing}.

\subsection{Problem formulation}
The scenario of a swarm of drones that perform sensing is formulated by considering the following performance metrics: (i) mission efficiency; (ii) sensing accuracy; and (iii) energy cost.

\cparagraph{Mission efficiency} It denotes the ratio of sensing values in all cells at all timeslots collected by the drones during their mission over the total required values in all cells at all timeslots during the period $t$. The purpose is to collect sensor data as much as possible. It is formulated as:
\begin{equation}
    \mathsf{Eff}(a^u, t) = \frac{\sum^{N}_{n=1}\sum^{S}_{s=1} v_{n,s}(a^u, t)}{\sum^{N}_{n=1}\sum^{S}_{s=1} V_{n,s}(t)}.
    \label{eq:effi}
\end{equation}

However, maximizing this metric leads to sensing imbalance and even blind areas. Some cells may be covered for a long time while some other cells may be never covered, i.e., the over-sensing and under-sensing~\cite{qin2024m}. Thus, it is necessary to consider matching indicators such as the sensing accuracy~\cite{qin2024m} to access such imbalances.

\cparagraph{Sensing accuracy} It denotes the matching (correlation\footnote{Error and correlation metrics (e.g. root mean squared error, cross-correlation or residuals of summed squares) estimate the matching, shown to be NP-hard combinatorial optimization problem in this context~\cite{pournaras2018decentralized}.}) between the total sensing values collected and the required ones. The metric is formulated as follows:
\begin{equation}
    \mathsf{Acc}(a^u, t) = \sqrt{ \frac{N \cdot S}{\sum^{N}_{n=1}\sum^{S}_{s=1}[v_{n,s}(a^u, t) - V_{n,s}(t)]^2} }.
    \label{eq:acc}
\end{equation}

Apart from improving the efficiency and accuracy of drone sensing, the energy consumed by drones needs to be saved.

\cparagraph{Energy cost} It is the battery usage of drones to perform sensing. A power consumption model~\cite{monwar2018optimized} is used to calculate the power consumption with input the specification of drones (weight, propeller and power efficiency) and the weather conditions (wind speed and temperature). This model estimates the cost of navigation and sensing plans, and emulates the outdoor environments~\cite{qin2024m}. The energy cost of each drone $u$ is formulated as:
\begin{equation}
    E(a^u ,t) = \frac{C^\mathsf{f}(u) \cdot t^\mathsf{f}(a^u) + C^\mathsf{h}(u) \cdot t^\mathsf{h}(a^u)}{B (u)},
    \label{eq:energy}
\end{equation}
where $B (u)$ is the battery capacity of drone $u$; $C^\mathsf{f}(u)$ and $C^\mathsf{h}(u)$ denote the flying and hovering power consumption of $u$ respectively; $t^\mathsf{f}$ and $t^\mathsf{h}$ are the flying and hovering time respectively, which are determined by the sensing plan $P^u(a^u, t)$. 

As described before, the optimization problem has three objectives: (i) maximize mission efficiency; (ii) maximize sensing accuracy; and (iii) minimize energy cost. Simultaneously achieving all of these objectives is a daunting task, as they often oppose each other. Maximizing both efficiency and accuracy is a trade-off. In addition, in order to achieve both of these goals, drones have to continuously move to collect data from different cells, which results in higher energy costs due to long-distance flights. However, some trips of drones may not improve sensing quality in the long run. For instance, at the end of a period, drones opt to return to the nearest charging station where there are fewer sensing tasks in the next period. This choice reduces the energy cost of the current period but comes at the significant expense of sacrificing the sensing quality in the next period. Therefore, the overall sensing performance is optimized by integrating these opposing objectives and selecting the plans of each drone. 

\cparagraph{Overall performance}
The objective function is defined as the weighted sum of mission efficiency, sensing accuracy and energy cost:
\begin{align}
        \mathop{\arg\max}\limits_{a^u, u \in \mathcal{U}} \sum^{U}_{u=1} \sum^{T}_{t=1} [\alpha_1 \mathsf{Eff}(a^u, t) + \alpha_2 \mathsf{Acc}(a^u, t) - \alpha_3 E(a^u, t)].
    \label{eq:objective}
\end{align}

\section{The Proposed Solution}

\begin{figure}[!t]
	\centering
	\includegraphics[width=12cm]{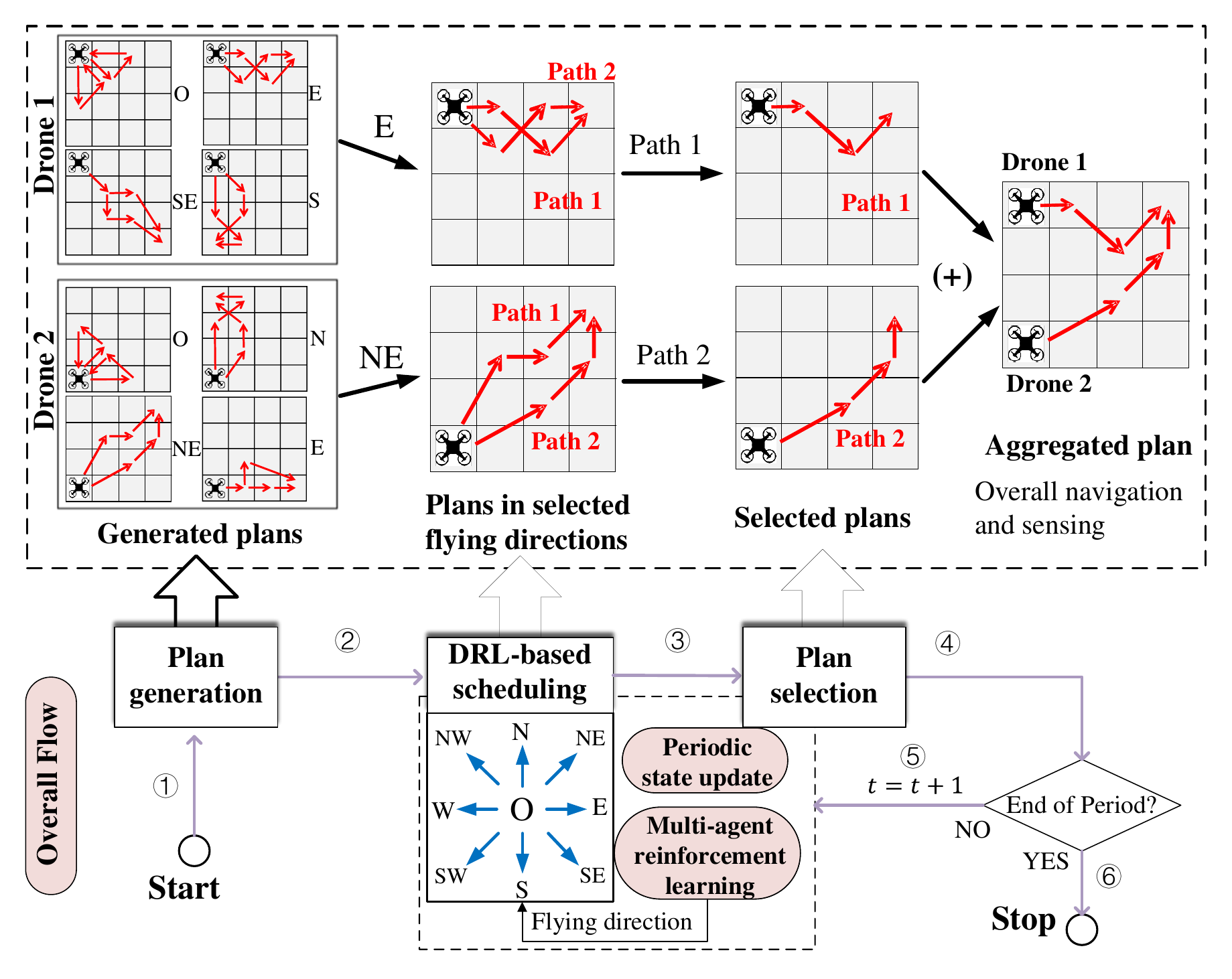}
	\caption{The hierarchical decision-making framework of \emph{DO-RL}. The overall flow of \emph{DO-RL} is depicted at the bottom; The plan selection is illustrated for two drones; The plan generation part outlines the procedure for generating a plan when drone $1$ is traveling to the east.}
	\label{fig:framework}
\end{figure}

In our problem, traditional DRL approaches suffer from the large action space and dimensional curse caused by the increasing number of time periods, leading to decline in sensing quality.
To ensure high-quality sensing performance, the proposed \emph{DO-RL} integrates decentralized drone coordination for short-term navigation and sensing optimization with long-term scheduling of flying directions.
Fig.\ref{fig:framework} illustrates the designed system framework of \emph{DO-RL}, consisting of three main components:

\begin{enumerate}
    \item \emph{Local plan generation} This component generates the navigation and sensing options for drones. Given the sensing map, each drone autonomously generates a finite number of discrete plans to initialize the overall process. This provides flexibility for the drones at the next stage to choose in a coordinated way. The generated plans are grouped into different flying directions. 
    \item \emph{Iterative plan selection} This distributed component leverages a collective learning to coordinate drones to locally select the optimal navigation and sensing options from their generated plans within a period. 
    \item \emph{DRL-based scheduling} This component is the core of the framework, which leverages the DRL algorithm to enable drones to take the actions of overall flying directions between departure and destination charging stations in each time period. Thus, these actions are executed period-by-period, with each made only after the drone completes its sensing missions in the current period. The component includes two elements: a periodic state update, which updates the state of drones for DRL-based scheduling in the next period, and a multi-agent reinforcement learning module, which is built based on centralized training and decentralized execution. 
\end{enumerate}

The \emph{DO-RL} process works as follows: Each drone initially generates a set of plans. It then selects a flying direction and chooses a subset of plans aligned with that direction. Each drone autonomously picks a plan from this subset, shares it with the swarm, and observes others' choices. Based on its action and observations, the drone calculates a reward function and updates its state (including location, battery level, and sensing requirements), and stores these results in a multi-agent reinforcement learning buffer to refine its flight strategies. This cycle of DRL-based scheduling and plan selection repeats at each time period until the mission is complete. However, \emph{DO-RL} requires frequent information sharing, which makes communication inefficient and vulnerable to potential failure of individual drones.

\begin{figure}[!t]
	\centering
	\includegraphics[width=8cm]{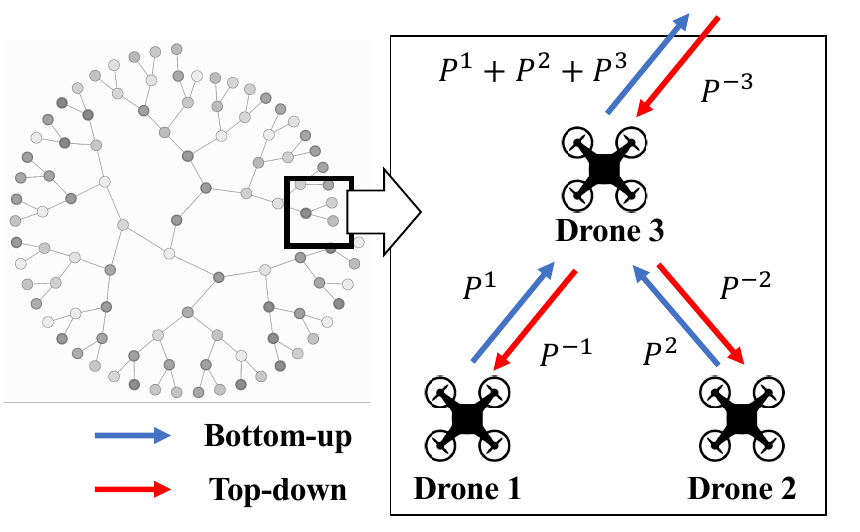}
	\caption{Overview of the tree communication topology. During the bottom-up process, Drone 3 aggregates the plans of its children, i.e., drone 1 and drone 2, and sends them to its parent agent with its own plan. During the top-down process, each parent agent sends the total aggregated plans to its children such that all agents obtain the observed plan.}
	\label{fig:tree_structure}
\end{figure}

To overcome these barriers, \emph{DO-RL} leverages a tree communication topology. Each drone, controlled by a local agent, locates and connects with other agents into a tree structure, as shown in Fig.~\ref{fig:tree_structure}. Once agents have established their proximity-based relationship, e.g., Euclidean distance, they are positioned starting from the leaves up to the root, each interacting with its children and parent in a bottom-up and top-down phase. 
For example, a drone $u$ aggregates and shares its plan $P^u$ in a bottom-up fashion, and then obtains the observed plan $P^{-u}$ in the top-down process~\cite{pournaras2018decentralized}. These plans serve as the observation of an agent in DRL modeling. As this coordination among agents using hierarchical organization reduces the need for every node to communicate with every other node, the system is highly-efficient in collaboration and information exchange~\cite{pournaras2018decentralized}. 

Through iterative bottom-up and top-down exchanges, agents only share the aggregated plans of navigation and sensing, rather than their sensitive information including the battery level and collected sensor data. This privacy-preservation~\cite{pournaras2018decentralized} is critical when the approach is applied to heterogeneous drones that belong to various entities, such as aviation agencies, private citizens, and companies. Moreover, drones perform the iterative collective learning process over a holarchic tree structure in response to communication failures, mitigating learning performance in case part of the network is disconnected. The resilience and privacy-preservation of this system has been previously demonstrated with large-scale real-world datasets~\cite{pournaras2018decentralized}, and thus this paper focuses on sensing performance and scalability of the proposed solution.


\section{Detailed System Design} 
This section illustrates the detailed system design of \emph{DO-RL}. We firstly model our problem as DRL, and then present the main components of \emph{DO-RL}, including the local plan generation and iterative plan selection. In addition, the periodic state update and the training of multi-agent reinforcement learning the DRL-based scheduling are introduced.

\subsection{DRL modeling} 
The core design of \emph{DO-RL} lies in applying DRL on our problem. Once drones take actions of flying directions, they execute plans and return to charging stations, changing their current locations and battery levels while observing the navigation and sensing of other drones. Therefore, the problem scenario can be explained into a Markov decision process~\cite{cui2019multi}. We model the problem using state, action, and reward concepts: 

\begin{enumerate}
    \item \emph{State}. The state $\textbf{s}_t$ at period $t$ consists of four components $(\mathcal{S}_1, \mathcal{S}_2, \mathcal{S}_3, \mathcal{S}_4)$, where $\mathcal{S}_1 = \{ c^u(t) |_{u \in \mathcal{U}} \}$ is the current locations of drones. Since drones charge at charging stations before taking actions, the location can serve as the index of the charging station. $\mathcal{S}_2 = \{ e^u(t) |_{u \in \mathcal{U}} \}$ is the current battery levels of drones, which are calculated based on the energy cost. $\mathcal{S}_3 = \{ P^u(t) |_{u \in \mathcal{U}} \}$ is the plan of $u$. $\mathcal{S}_4 = \{ P^{-u}(t) |_{u \in \mathcal{U}} \}$ is the aggregated plan of other drones excluding $u$, which are shared via the tree structure. 
    \item \emph{Action}. The action $\textbf{a}_t = \{ a^1_t,...,a^U_t \}$ at period $t$ consists of period-by-period flying directions (N, E, S, W, NE, SE, SW, NW or O) determined by drones. 
    \item \emph{Reward}. Based on the objective function of Eq.(\ref{eq:objective}), the expected immediate local reward of one drone at period $t$ is defined as follows: $\textbf{r}^u_t = \alpha_1 \mathsf{Eff}(a^u, t) + \alpha_2 \cdot \mathsf{Acc}(a^u, t) - \alpha_3 E(a^u, t)$. Throughout the training process or episodes, the overall reward fluctuates based on the actions taken by the drones. This helps drones in prioritizing areas rich in sensor data by maximizing their reward, i.e., the highest cumulative overall performance.
\end{enumerate}

\begin{figure}[!t]
	\centering
	\includegraphics[width=10cm]{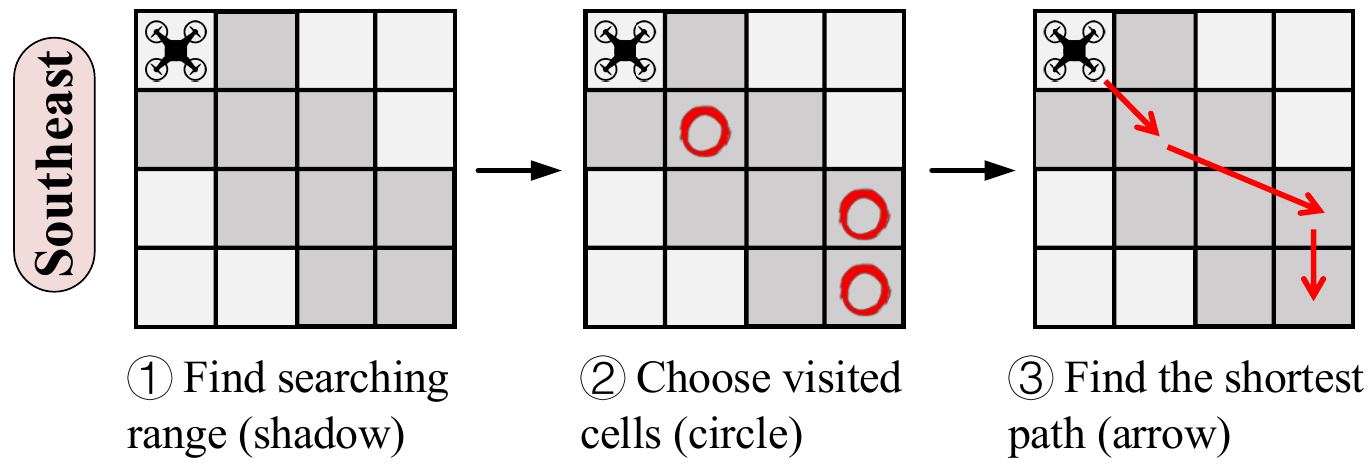}
	\caption{Process of path finding in plan generation.}
	\label{fig:plan_generation}
\end{figure}

\subsection{Local plan generation}
The actions enable each drone to generate nine groups of plans, each associated with an action for a flying direction. These groups consist of multiple individual plans of the same size. A group of plans generated by drone $u$ during a period is defined as $G^{a^u}(t)$, where each plan in the group has the same action, $P^u(a^u, t) \in G^{a^u}(t)$. Then, a plan can be defined as $P^u(a^u, t) = \{ l, K(a^u), J(a^u), E(a^u, t) \}$, where $l$ is the index of the plan, $l \leq L$; $K(a^u)$ denotes the indexes of cells within the searching range along flying direction $a^u$; $J(a^u)$ denotes the indexes of visited cells within $K(a^u)$; and $E(a^u, t)$ denotes the energy consumed by $u$ traveling over $J(a^u)$, i.e., the cost of the plan.

The details of generating a single plan are defined by the chosen path and the corresponding energy required to traverse that path, as shown in Fig.~\ref{fig:plan_generation}. As an example of a plan, the drone $u$ firstly finds a searching range to cells $K(a^u)$: it determines a range with a certain width (equal to half of distance to nearest charging station) along its flying direction, and then searches the cells within this range. Next, it selects the visited cells $J(a^u)$ randomly from this range. Then, $u$ finds the shortest possible path among $J(a^u)$ via the \textit{Dijkstra's algorithm} for the Traveling Salesmen Problem. The path includes the information of which cell and which timeslot the drone uses to collect data, and thus both the traveling time $t^\mathsf{f}(a^u)$ and sensing time $t^\mathsf{h}(a^u)$ are calculated. The total energy consumption $E(a^u, t)$ when $u$ travels over the path is then calculated via Eq.(\ref{eq:energy}).

\subsection{Iterative plan selection}
Given the plans in selected flying directions within a period, drones can adapt their allocations by selecting appropriate plans in response to changes in task targets. In this process, the agents of drones improve their plan selections via tree communication.
Specifically, each agent obtains the aggregated choices, i.e., the aggregated plan $P^{-u}(a^u, t)$, from other agents, and chooses one of the plans $P^u(a^u, t, l)$ such that all choices together $P^u(a^u, t) + P^{-u}(a^u, t)$ add up to match a target $\mathcal{R}(t)$. The target is used to steer each drone to sense over a unique cell during a timeslot, thereby avoiding the simultaneous sensing of multiple drones within the same cell, i.e., preventing over-sensing and under-sensing.

The purpose of coordination is to match the aggregated plans of all agents to the target while minimizing the energy cost $E(a^u, t)$ of the plan selected by the drone. The cost function for each agent is formulated using the root mean square error (RMSE):
\begin{align}
    \mathop{\min}\limits_{a^u, u \in \mathcal{U}} (1 - \beta) \cdot \sqrt{ \frac{ \sum^N_{n=1} \sum^S_{s=1} [p_{n,s}(a^u, t) - r_{n,s}(t)]^2 }{ N \cdot S } } + \beta \cdot E(a^u, t),
    \label{eq:epos}
\end{align}
where $\beta$ represents the behavior of an agent. As the value of $\beta$ increases, the agent becomes increasingly selfish, prioritizing plans with lower energy costs at the expense of higher root mean squared error. Each agent (or drone) aims to minimize the system-wide cost of sensing while considering its energy consumption. 
By minimizing this cost function, each drone coordinates to select its optimal plan, indicating a specific path within a period (comprising multiple timeslots). Via planning and selection, drones using \emph{DO-RL} only need to take high-level actions of flying directions in each period, and thus reducing the number of decisions and observations in DRL.

\subsection{Periodic state update}\label{sec:update}
After choosing a plan, each drone changes its current state at the next period $t + 1$, including its location $c^u(t+1)$, battery level $e^u(t+1)$, selected plan $P^u(t+1) := P^u(a^u, t)$, and observed plan $P^{-u}(t+1) := P^{-u}(a^u, t)$. However, due to dynamic changing environment, drones have no knowledge about the required sensing value in the next period. They need to predict and estimate the required sensing value to calculate reward function via Eq.(\ref{eq:effi}) and (\ref{eq:acc}). Depending on the predicted distribution of sensor data, the target is required to be updated.

The predicted sensing value at period $t$, denoted as $\hat{V}(t)$, is updated in a time-reverse decay, formulated as follows:
\begin{equation}
    \hat{V}_{n,s}(t) =\sum^t_{t'=1}(T - t + t') \cdot \omega_{n,s}(t') \cdot v'_{n,s}(a^u, t') := V_{n,s}(t),
    \label{eq:v_history}
\end{equation}
where $V'_{n,s}$ denotes the data values collected by drones once they execute their selected plans; $\omega_{n,s}(t')$ is a prediction coefficient such that $0 < \omega_{n,s}(t') < 1$, $t' \leq t$. To achieve accurate predictions, \emph{DO-RL} leverages the Ordinary Least Squares regression method (OLS) to train these coefficients initially, and use the past experienced observations as the target distribution for training. 

The target $\mathcal{R}$ in the plan selection needs to be iteratively updated to steer the coordination of drones to choose plans for areas and timeslots with abundant sensor data. The percentile-based data filtering is used to eliminate the extreme low sensing values collected by drones. This approach effectively removes the need for data collection in regions with low sensing requirements (e.g., traffic exclusion zones). As a consequence, it helps drones in prioritizing their operations in cells and time where sensing values are significantly high. We initialize the target as $r_{n,s}(0)=1$ to encompass all cells and timeslots, and update it as follows:
\begin{equation}
    \begin{split}
        r_{n,s}(t) = \left \{
        \begin{array}{ll}
           0,       &    v'_{n,s}(a^u, t) < \overline{v} \wedge p_{n,s}(a^u, t) > 0\\
           r_{n,s}(t-1), &    otherwise
        \end{array},
        \right.
\end{split}
\label{eq:target}
\end{equation}
where the threshold $\overline{v}$ is set iteratively and calculated as the value at the $100(1 - U/N)$th percentile among the predicted data values $\hat{V}_{n,s}(t)$. 

At the end of state update in \emph{DO-RL}, the current collected data values $V'_{n,s}$, the predicted values $\hat{V}_{n,s}$ and the target $\mathcal{R}$ are shared with each agent via the top-down interactions within the tree communication structure. This information will be stored and later sampled for multi-agent reinforcement learning.

\begin{algorithm}[t]
    \caption{\textit{DO-RL} training}
    \label{algorithm1}
    Initialize critic network $Q(\cdot)$, actor network $\pi(\cdot)$ with weights $\theta^Q$, $\theta^\pi$, and their two target networks $Q'(\cdot)$, $\pi'(\cdot)$ with parameters $\theta^{Q'} := \theta^Q $, $\theta^{\pi'} := \theta^\pi$\;
    \For{episode $:= 1$ to max-episode-number}
    {
        Reset environment and obtain the initial state $\textbf{s}_1$\;
        \For{period $t := 1$ to episode-length}
        {
            \For{drone $u := 1,...,U$}
            {
                Take action $a^u _t = \pi(\textbf{o}^u_t | \theta^\pi)$\;
                Select the set of plans $G^{a^u}(t)$\;
                Pick a plan $P^u(a^u, t)$ and share it with other drones via tree communication\; 
                Observe the aggregated plan $P^{-u}(a^u, t)$ and pick an optimized plan through collective learning~\cite{pournaras2018decentralized}\;
                Compute reward $\textbf{r}^u_t = \alpha_1 \mathsf{Eff}(a^u, t) + \alpha_2 \cdot \mathsf{Acc}(a^u, t) - \alpha_3 E(a^u, t)$\;
                Store transition sample $(\textbf{o}^u_t, \textbf{a}^u_t, \textbf{r}^u_t, \textbf{o}^u_{t+1})$ into buffer\;
                Sample a random minibatch of $H$ samples $(\textbf{o}^u_k, \textbf{a}^u_k, \textbf{r}^u_k, \textbf{o}^u_{k+1})$ from buffer\;
            }
        }
        Update $\theta^\pi$ by minimizing the actor loss function via Eq.(\ref{eq:actor})\;
        Update $\theta^Q$ by minimizing the critic loss function via Eq.(\ref{eq:critic})\;
        Update the target networks: $\theta^{Q'} \leftarrow \theta^Q $, $\theta^{\pi'} \leftarrow \theta^\pi$\;
    }
\end{algorithm}

\begin{figure}[!t]
	\centering
	\includegraphics[width=8cm]{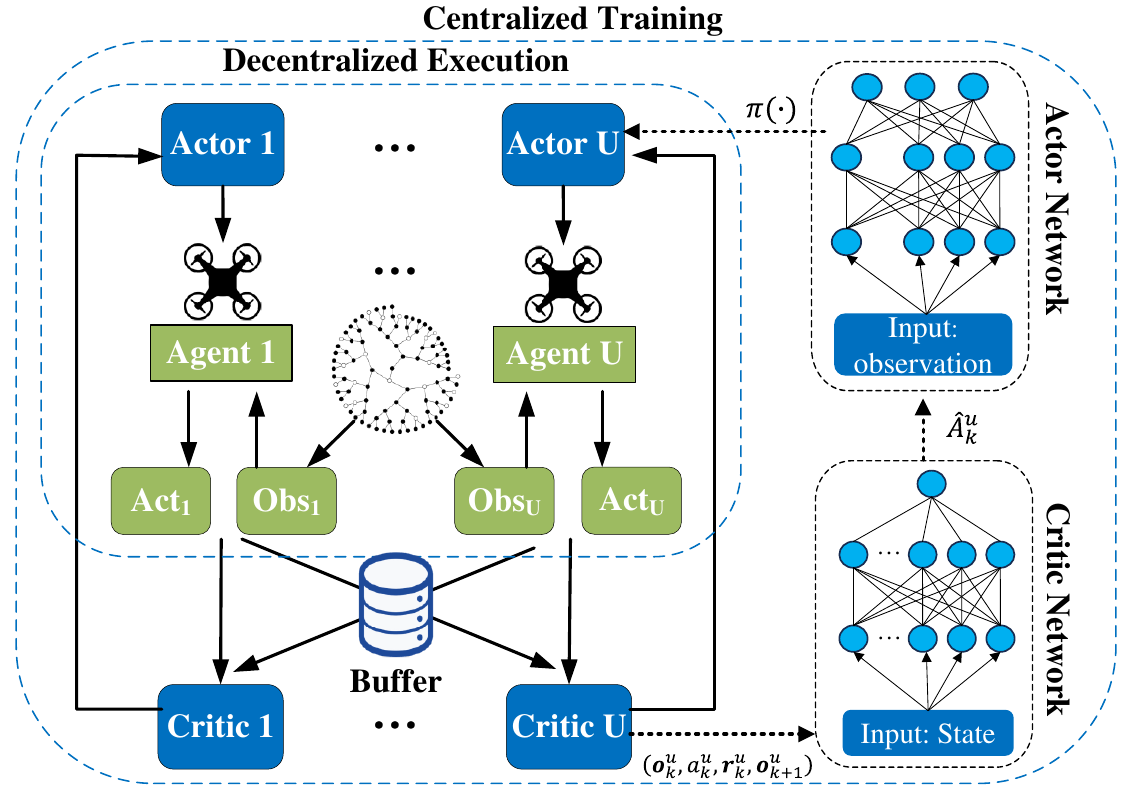}
	\caption{DRL-based long-term scheduling overview.}
	\label{fig:drl}
\end{figure}

\subsection{Multi-agent reinforcement learning}\label{sec:DRL}
Our approach \emph{DO-RL} provides a generalized model for multi-agent reinforcement learning by adopting the framework of centralized training and decentralized execution, as shown in Fig.\ref{fig:drl}. During execution, each agent obtains a local aggregated observation comprising public information. Apart from their own private data such as locations, battery levels, and selected plans, this observation contains aggregated plans shared via the tree communication topology (shown in Fig.~\ref{fig:tree_structure}). 
The training part is built based on an actor-critic model. The actors allow drones to work independently and choose actions based on their observations and policies, allowing for parallel data collection and processing. The critic, however, evaluates the actions of drones and guide them toward collectively optimal decisions. In addition, \emph{DO-RL} employs Proximal Policy Optimization (PPO) to prevent detrimental updates and improving the stability of the learning process~\cite{yi2022automated}. Algorithm~\ref{algorithm1} illustrates the overall training process of \emph{DO-RL}.

The process of training primarily involves the following steps: Firstly, each agent takes an action for a flying direction based on the aggregated observation $\textbf{o}^u_t$ shared via coordination as well as the reward $\textbf{r}^u_t$. Once all actions are determined, drones select their plans and transition to a new state. Subsequently, the buffer, a data storage structure used for experience replay, stores all transitions of each agent $(\textbf{o}^u_t, \textbf{a}^u_t, \textbf{r}^u_t, \textbf{o}^u_{t+1})$. Several groups of transitions $(\textbf{o}^u_k, \textbf{a}^u_k, \textbf{r}^u_k, \textbf{o}^u_{k+1})$ are sampled randomly ($H$ groups of transitions) for updating the parameters of both the critic and actor networks. The algorithm is typically an extension of the actor-critic policy gradient approach, employing two deep neural networks for each agent: a critic network $Q(\cdot)$ and an actor network $\pi(\cdot)$. The critic network estimates the reward associated with a transition using Bellman equation, and its parameter $\theta^Q$ is then updated by minimizing a critic loss function $L(\theta^Q)$:
\begin{equation}
    L(\theta^Q) = \frac{1}{H \cdot U} \sum^H_{k=1} \sum^U_{u=1} (\hat{A}^u_k)^2, 
    \label{eq:critic}
\end{equation}
\begin{equation}
    \hat{A}^u_k = \textbf{r}^u_k + \gamma \cdot Q(\textbf{o}^u_{k+1}, a^u_{k+1}) - Q(\textbf{o}^u_{k}, a^u_{k}),
\end{equation}
where $\gamma$ is a discount factor; $\hat{A}^u_k$ is an advantage function.
Then, the critic network provides $\hat{A}^u_k$ to the actor network to increase the probability of actions that have a positive impact and decrease the ones that have negative impact. The actions are taken by drones as $a^u _t = \pi(\textbf{o}^u_t)$. The parameter $\theta^\pi$ of the actor network is updated by maximizing the clip objective, or minimizing the actor loss function $L_{CLIP}(\theta^\pi)$, as follows~\cite{yi2022automated}:
\begin{align}
    \begin{split}
        L_{CLIP}(\theta^\pi) =  \;
        & \frac{1}{H \cdot U} \sum^H_{k=1} \sum^U_{u=1} min(ratio(\theta^\pi, u)\hat{A}^u_k, \\
        & clip(ratio(\theta^\pi, u), 1 - \epsilon, 1 + \epsilon)\hat{A}^u_k), 
    \end{split}
    \label{eq:actor}
\end{align} 
where $\epsilon$ is a hyperparameter; $clip(\cdot)$ defines the surrogate objective by limiting the range of $ratio(\theta^\pi, u) = \frac{\pi(a^u_k | \textbf{o}^u_k)}{\pi_{old}(a^u_k | \textbf{o}^u_k)}$ using clipping, thereby eliminating incentives to exceed the interval $[1 - \epsilon, 1 + \epsilon]$; $\pi_{old}$ denotes the older policy of the actor network in the previous iteration. 

\section{Performance Evaluation}
In this section, an overview of the experimental settings is presented. The sensor dataset, specification of drones, optimization algorithm employed, and the used neural network are introduced. Then, the baselines and performance evaluation metrics are discussed. Finally, the results are assessed across various scenarios.

\begin{figure}[!t]
    \centering
    \includegraphics[width=9cm]{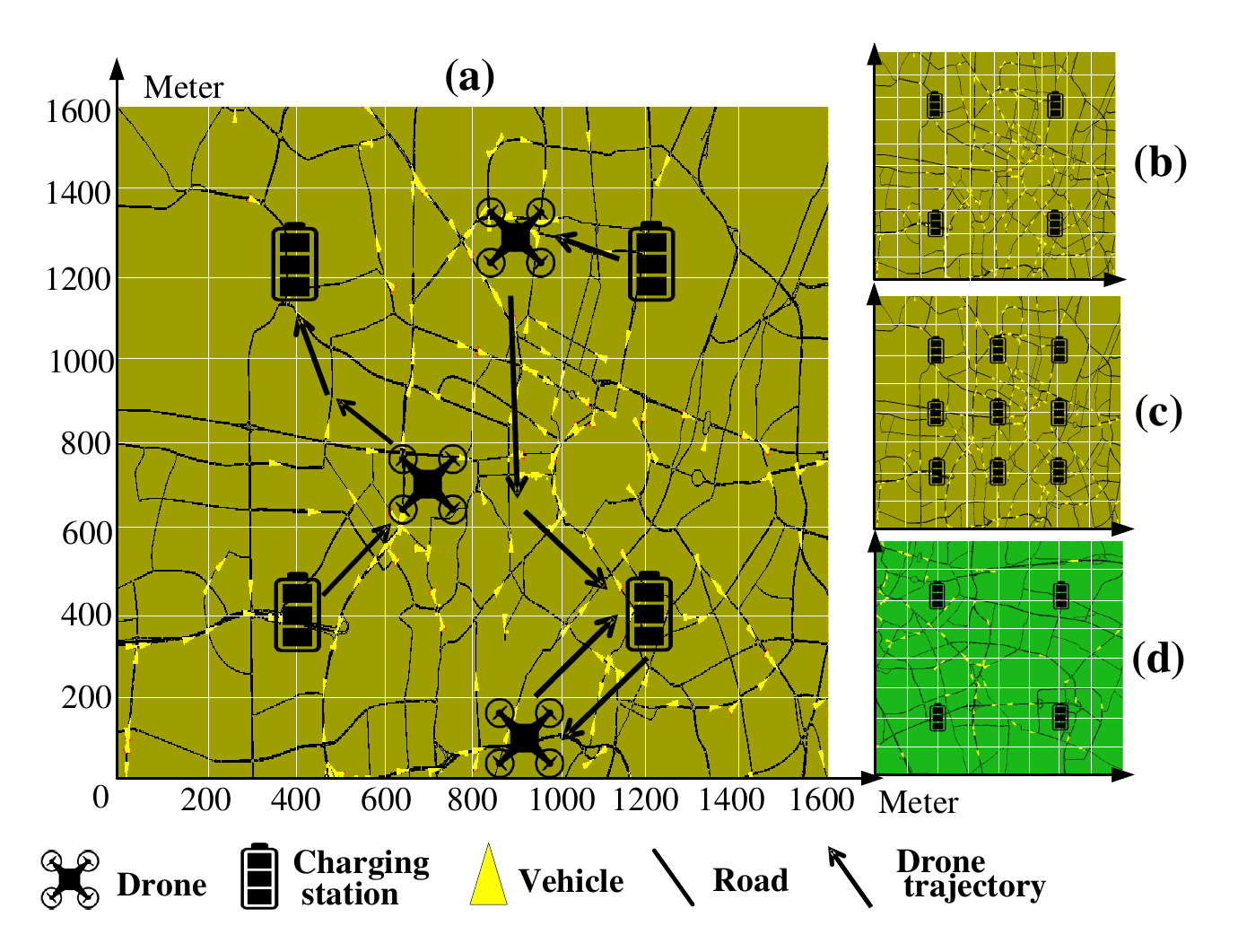}
    \caption{The central business district of Munich, Germany: (a) Basic scenario with $64$ cells, $4$ charging stations and high density of vehicles; (b) Increase the number of cells to 100; (c) Increase the number of charging stations to 9; (d) Change to a new map with low density of vehicles.}
    \label{fig:munich}
\end{figure}

\begin{figure}[!t]
    \centering
    \includegraphics[width=8cm]{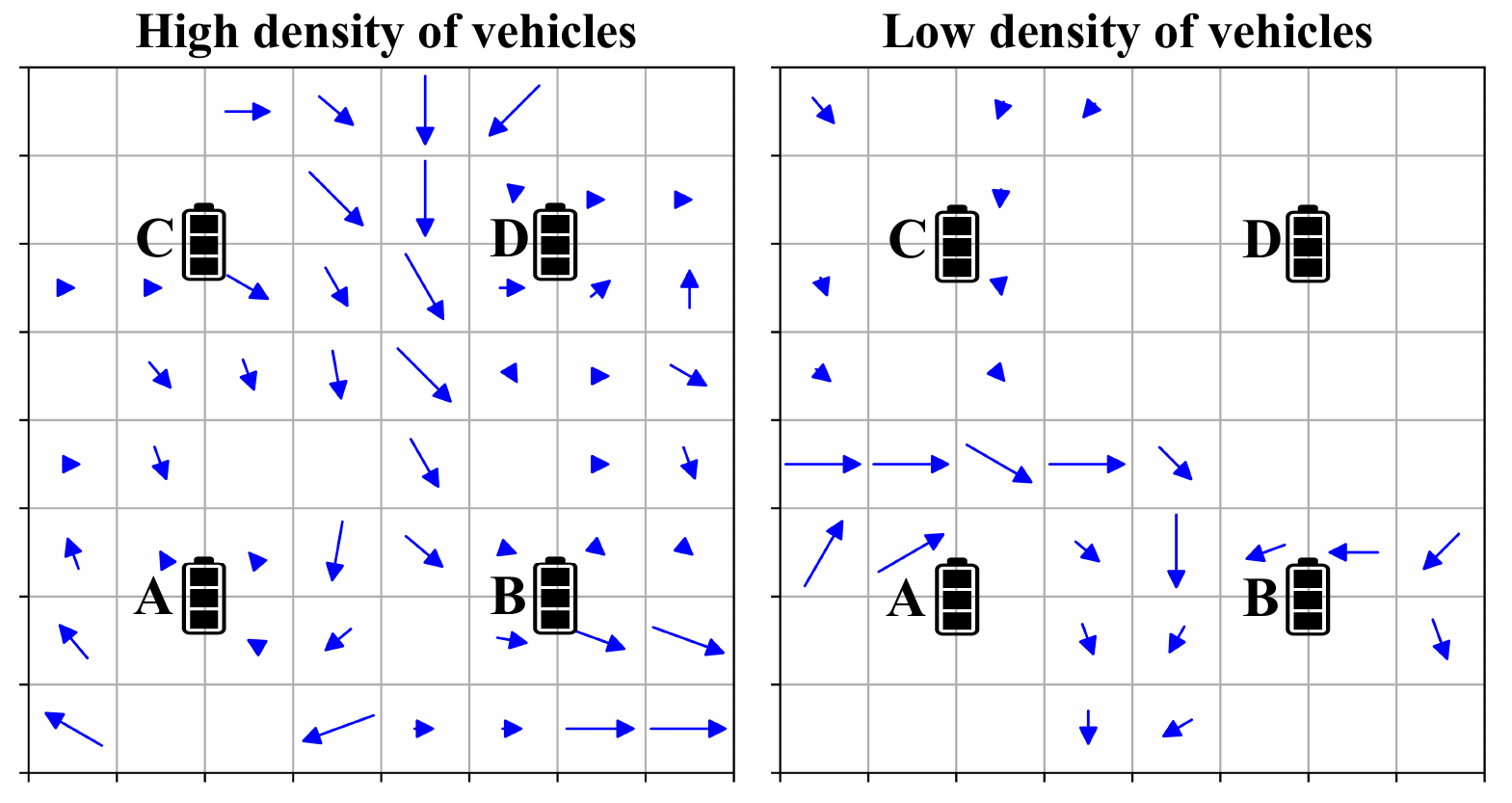}
    \caption{The distribution of both charging stations and traffic vehicles in the maps with high and low density of vehicles. There are $4$ charging stations uniformly distributed in the map with $64 = 8 \times 8$ cells lined up over the map. The blue arrows symbolize the flow of vehicular traffic, with their length proportional to the volume of vehicles over $8$ periods.}
    \label{fig:charging_dist}
\end{figure}

\subsection{Experimental settings} \label{sec:ex_settings}
In order to evaluate the \emph{DO-RL}, a real-world transportation scenario is modeled, where a swarm of drones perform the sensing tasks of traffic monitoring. The number of vehicles serves as the required sensing values in the model. The experimental scenario is based on a detailed traffic flow of Munich, Germany. A realistic traffic distribution is generated using real-world traffic flow data\footnote{Munich IoT Benchmarking Dataset, available at: https://github.com/Znbne/MunichIoT.} representing actual vehicle journeys within the city~\cite{nezami2025computing}, integrated with the SUMO (Simulation of Urban Mobility) traffic simulator\footnote{SUMO, available at: https://www.eclipse.org/sumo/.}. To model dynamic traffic behavior and improve the realism of vehicle routing, the duaIterate algorithm\footnote{Duarouter, available at: https://sumo.dlr.de/docs/duarouter.html.} is employed to iteratively refine the vehicle routes to optimize traffic flow and mitigate congestion~\cite{nezami2025computing}. 

Fig.~\ref{fig:munich} illustrates a selected map of $1600 \times 1600$ meters in the city with the simulation time of $50$ hours, which is $100$ periods. It has a high density of vehicles, approximately $2,000$ vehicles passing by per hour. The area split into a finite number of cells, each is defined as a rectangular square with the size of $200 \times 200$ meters that can be captured by the cameras of drones. Fig.~\ref{fig:charging_dist} shows the distribution of both charging stations and traffic vehicles in the map. The charging stations are uniformly distributed in the map. The cross-validation is employed: $80\%$ simulation time of the datasets for training and $20\%$ for testing. 

In terms of the drone, the DJI Phantom 4 Pro model\footnote{The specifications of DJI Phantom 4 Pro is available at: https://www.dji.com/uk/phantom-4-pro/info.} is considered. Each drone has a body weight of $1.07$ kg and four propellers, each with a diameter of $0.35$ m. The ground speed of the drones is set at $6.94$ m/s, and the drag force is determined to be $4.1134$ N~\cite{monwar2018optimized}. As a consequence, the power consumption remains consistent across all drones. Additionally, drones are equipped with a $6000$ mAh LiPo 2S battery ($0.31$ kg). The maximum flying time for drones is approximately $30$ min, which is set as one flying period. A flying period has $S = 30$ timeslots, each of one minute. Moreover, to ensure that the camera covers the entire area of a cell (see Fig.~\ref{fig:munich}), the minimum hovering height of drones is determined at which the field of view of the camera and the cell overlap. The hovering height $h$ is calculated based on the distance between any two cells (approximately $200$ meters), the camera resolution, focal length derived from camera calibration, and ground sampling distance~\cite{wierzbicki2018multi}. Therefore, each drone equipped with a 4K camera senses from a minimum height of $82.4$ meters.

Furthermore, the algorithm settings of \emph{DO-RL} is determined empirically regarding parameter selection for the evaluation. It contains the settings of distributed optimization and reinforcement learning. More results of the effect of different parameters are illustrated in Appendix~\ref{appendix_param}.

\cparagraph{Planning and distributed optimization}
The proposed \emph{DO-RL} leverages the decentralized multi-agent collective learning method of \emph{EPOS}\footnote{EPOS is open-source and available at: https://github.com/epournaras/EPOS.}, the \textit{Economic Planning and Optimized Selections}~\cite{pournaras2018decentralized}. Each agent, which is mapped to a drone, generates $L = 64$ plans, which are made openly available to encourage further research on coordinated sensing of drones~\cite{qin2024m}. Each plan indicates that a drone visits $2$ cells for sensor data collection during a trip. During the coordinated plan selection via \emph{EPOS}, agents self-organize into a balanced binary tree as a way of structuring efficient learning interactions. The shared goal of the agents is to minimize the RMSE between the total sensing values collected and the target via Eq.(\ref{eq:epos}). In the one execution of \emph{EPOS}, the agents perform $40$ bottom-up and top-down learning iterations during which RMSE converges to a minimum optimized value. In addition, drones are allowed to choose the their behavior within the range $\beta \in [0.1, 0.8]$.

\cparagraph{Neural network and learning algorithm} 
In this paper, the weight parameters $\alpha_1$, $\alpha_2$ and $\alpha_3$ are all set to a value of 1. A total of $H = 64$ transitions are sampled as a batch in a replay buffer, with a discount factor of $\gamma=0.95$ and a clip interval hyperparameter of $\epsilon=0.2$ for policy updating. We try the recurrent neural network (RNN), and use $W=64$ neurons in the two hidden layers of the RNN in both critic and actor networks. The activation function used for the networks is tanh. The models are trained over $\mathcal{E}=5000$ episodes, each consisting of multiple epochs (equal to the number of time periods).

\subsection{Baselines and metrics}
A fair comparison of the proposed method, \emph{DO-RL}, with related work is not straightforward as there is a very limited number of relevant algorithms. As discussed in Section~\ref{sec:related}, the traditional DRL (e.g., PPO) and distributed optimization methods (PSO, genetic) cannot be directly applicable to the decentralized combinatorial optimization problem defined in Section~\ref{sec:model}. Those relevant algorithms (e.g., COHDA, CBBA) have higher computation and communication overhead than the collective learning used in \emph{DO-RL}, which is illustrated in Section~\ref{sec:results}. For this reason, this section focuses on the ablation studies that removes one of the components of \emph{DO-RL}, including coordination, reinforcement learning and collective learning. The state-of-the-art baseline methods are listed as follows:
\begin{enumerate}
    \item \emph{Greedy}. It aims to assign each drone to a single cell, enabling it to accomplish the necessary sensing tasks within a time period~\cite{bartolini2019task}. This algorithm minimizes the number of visited cells and travel distance, resulting in reduced energy consumption for drones. To promote decentralization, we implement the algorithm with a local view for each drone, i.e., they have no coordination or awareness of cell occupancy by other drones. This approach ensures an energy-efficient and independent operation of each drone.
    \item \emph{EPOS}. It is designed to select the optimal plan for drones based on the generated plans~\cite{pournaras2018decentralized}. It incorporates the plan generation and the periodic state update, but does not support any long-term strategic navigation. Agents randomly choose any charging station to land on at every period. 
    \item \emph{MAPPO}. It is one state-of-the-art DRL algorithm using PPO~\cite{ding2021crowdsourcing,yi2022automated}, but does not include distributed optimization compared to \emph{DO-RL}. Moreover, agents in \emph{MAPPO} learn the flying directions timeslot by timeslot, rather than period by period as in \emph{DO-RL}. At each timeslot, a drone takes actions to horizontally move to an adjacent cell in eight directions or hover (similar to the actions in \emph{DO-RL}), and returns to the nearest charging station at the end of every $S$ timeslots (one period). For fair comparison, this method employs the same reward function and tree communication topology to share the aggregated observation.
\end{enumerate}

The evaluation of all algorithms includes three key metrics: (i) \emph{Mission efficiency}, (ii) \emph{Sensing accuracy} and (iii) \emph{Energy cost}. These metrics are determined using Eq.(\ref{eq:effi}), (\ref{eq:acc}) and (\ref{eq:energy}) respectively. We use the number of vehicles detected/observed by drones to represent the sensing values defined in this paper. Furthermore, to obtain a comprehensive assessment that considers all three metrics, the overall performance evaluation is conducted using Eq.(\ref{eq:objective}).

\begin{table}[!t]
	\centering
	\caption{Comparison of computational and communication costs.}  
	\label{table:complexity}
    \resizebox{\linewidth}{!}
    {
    \begin{tabular}{lcccccc}  
		\toprule  
		\textbf{Attributes \, Approaches.:} &\tabincell{l}{\emph{EPOS}~\cite{pournaras2018decentralized}} &\emph{COHDA}~\cite{hinrichs2013cohda} &\emph{CBBA}~\cite{bertuccelli2009real}
        &\tabincell{l}{\emph{MAPPO}~\cite{ding2021crowdsourcing}} & \emph{DO-RL}\\  
		\midrule     
            Computational Cost     &$O(T L I \log U)$   &$O(T L I)$     &$O(T L I U)$   &$O(\mathcal{E}TSUC_{dnn})$   &$O(\mathcal{E}T(U C_{dnn}+LI \log U))$\\ 
            Communication Cost    &$O(T I \log U)$  &$O(T I U)$    &$O(T I U)$   &$O(\mathcal{E} T U ^ 2)$   &$O(\mathcal{E} T I \log U)$\\ 
  
		\bottomrule
	\end{tabular}  
    }
\end{table} 

\subsection{Results and analysis} \label{sec:results}
Before evaluate the metric performance, the complexity of all approaches are compared.
Given the number of nodes per layer $W$, number of episodes $\mathcal{E}$, size of state space $\mathcal{X}$, and number of actions $A$, the computational complexity of DNN is approximately $O(C_{dnn}):=O((\mathcal{X}+A)W+3W^2)$~\cite{omoniwa2023communication}. The comparison of both computational and communication cost is shown in Table~\ref{table:complexity}, where $I$ denotes the number of iterations in \emph{EPOS}. 
The results illustrate that \emph{DO-RL} significantly lowers the computation cost by reducing the number of actions required during training on a temporal scale. In contrast, \emph{MAPPO} has higher action-state space, thereby leading to higher training complexity and exploration inefficiency in training. \emph{DO-RL} also has low communication cost due to the tree communication topology compared to other distributed optimization methods such as COHDA and CBBA. More results of complexity comparison and training convergence are shown in Appendix~\ref{appendix_compare}.

\begin{figure}[!t]
	\centering
        \subfigure{
		\includegraphics[height=3.2cm,width=3.8cm]{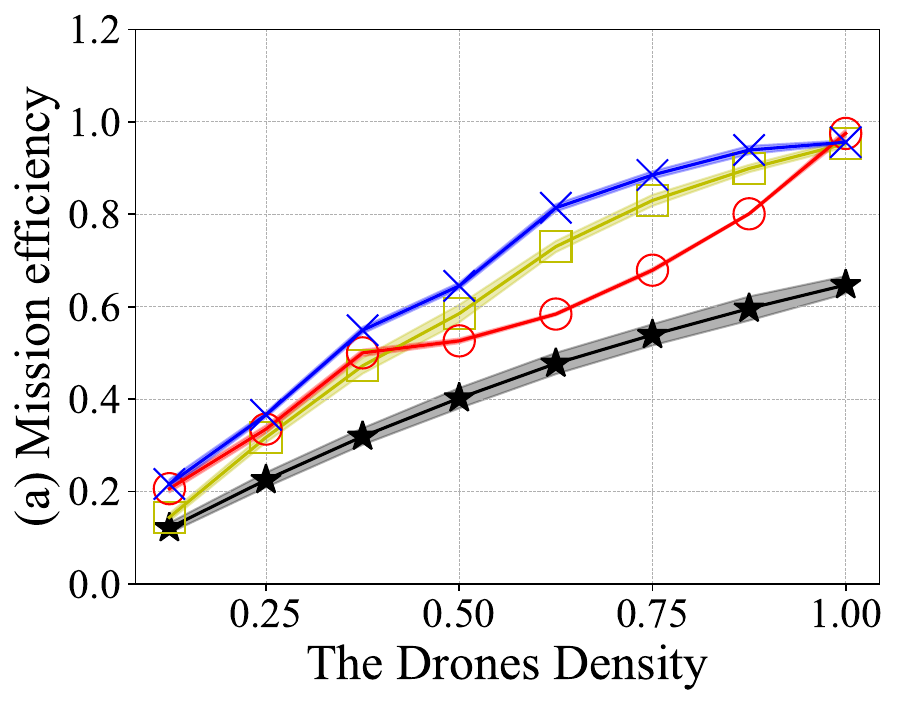}
		\label{fig:drones_effi}
	}
         \subfigure{
		\includegraphics[height=3.2cm,width=3.8cm]{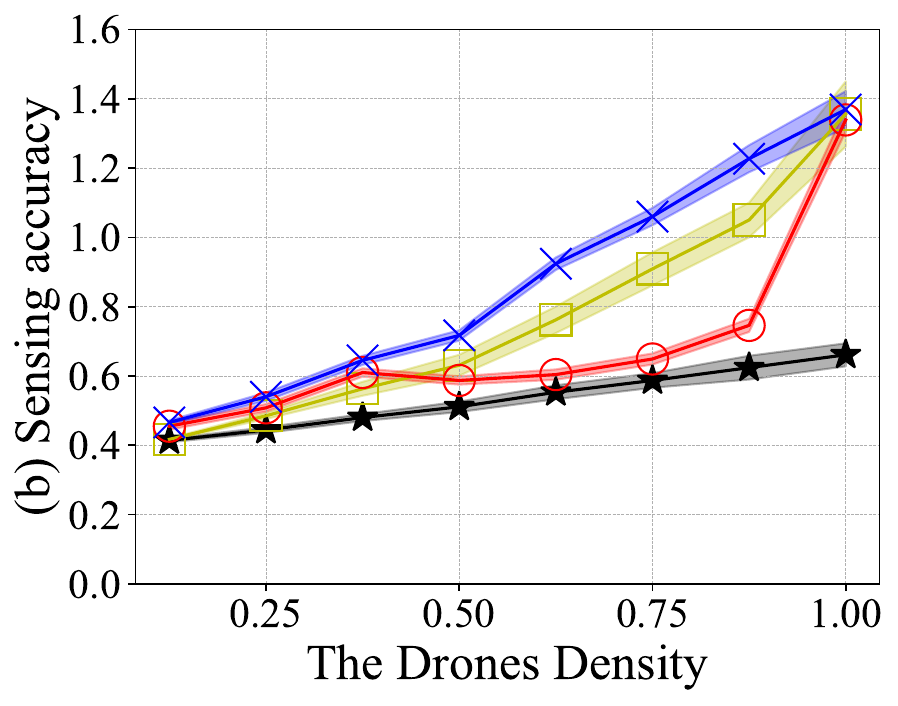}
		\label{fig:drones_acc}
	}
         \subfigure{
		\includegraphics[height=3.2cm,width=3.8cm]{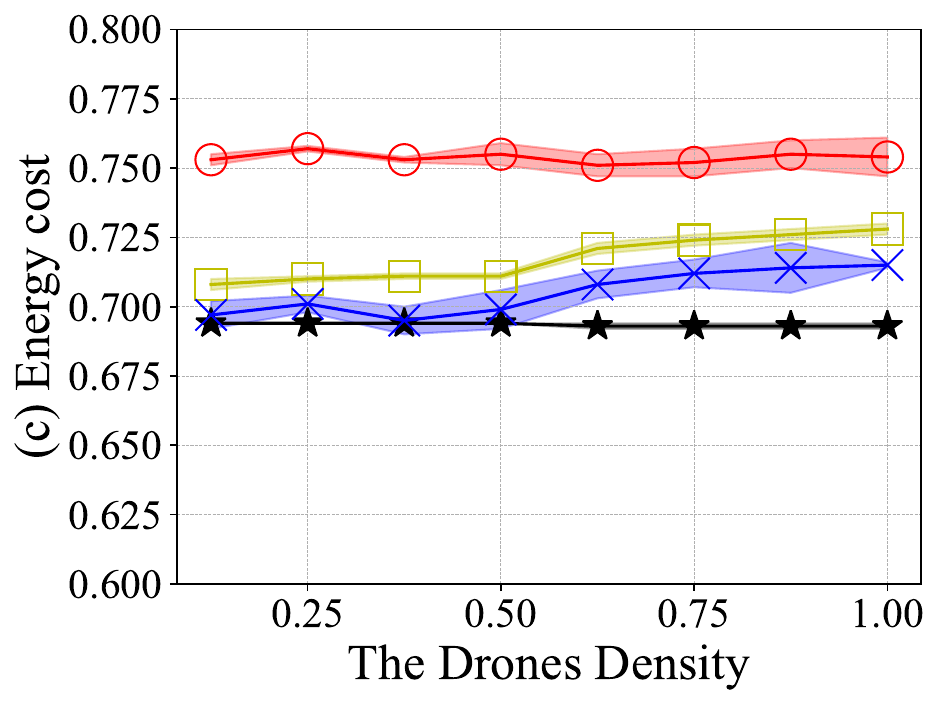}
		\label{fig:drones_energy}
	}
         \subfigure{
		\includegraphics[height=3.2cm,width=3.8cm]{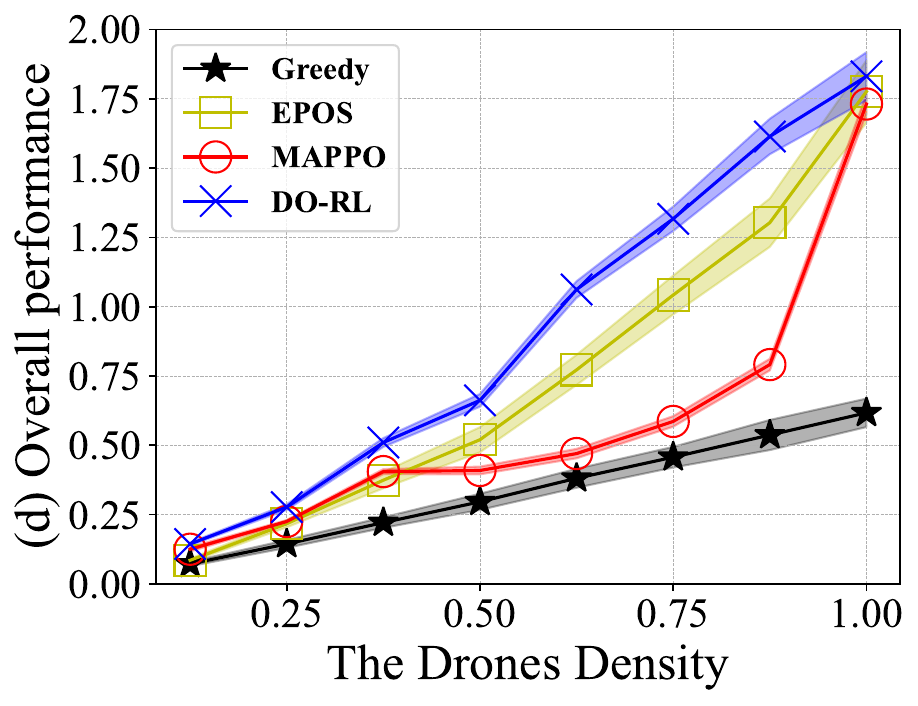}
		\label{fig:drones_overall}
	}
    \caption{\textbf{High density of drones increases both mission efficiency and sensing accuracy of all methods, especially for short-term optimization methods (\emph{DO-RL} and \emph{EPOS}).} DO-RL also shows its superior performance across different drones densities. Changing the drones density by increasing the number of drones from $8$ to $64$ and fixing 8 periods, 64 cells, 4 charging stations and high density of vehicles.}
    \label{fig:drones}
\end{figure}

\begin{figure}[!t]
	\centering
        \subfigure{
		\includegraphics[height=3.2cm,width=3.8cm]{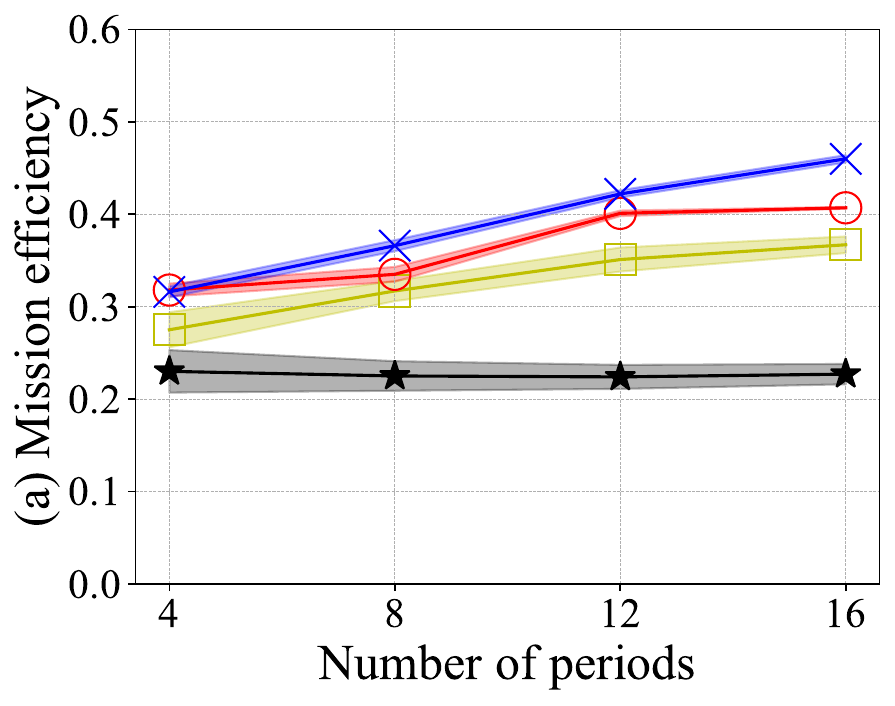}
		\label{fig:periods_effi}
	}
         \subfigure{
		\includegraphics[height=3.2cm,width=3.8cm]{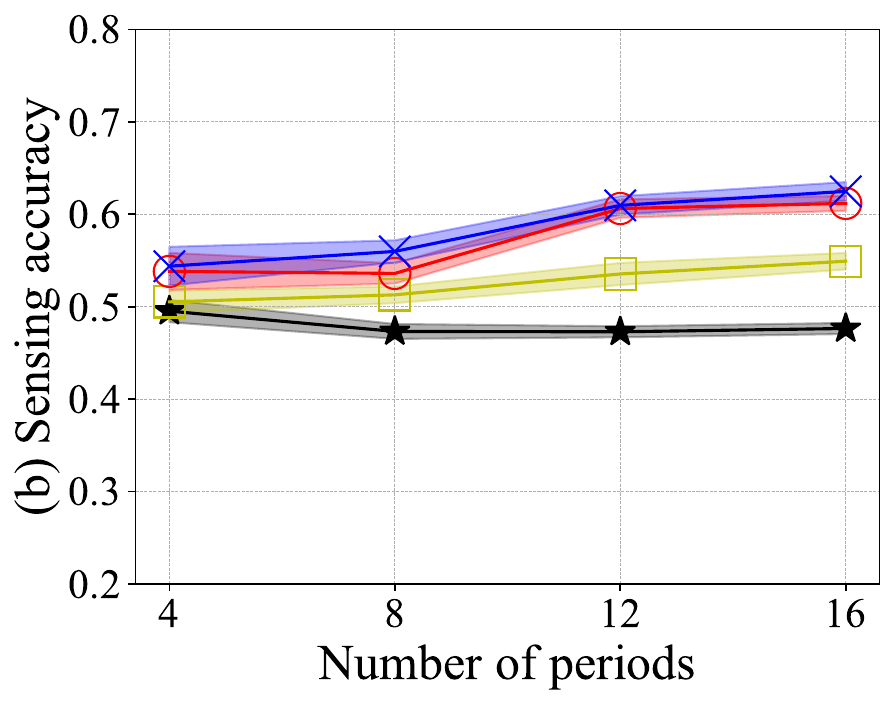}
		\label{fig:periods_acc}
	}
         \subfigure{
		\includegraphics[height=3.2cm,width=3.8cm]{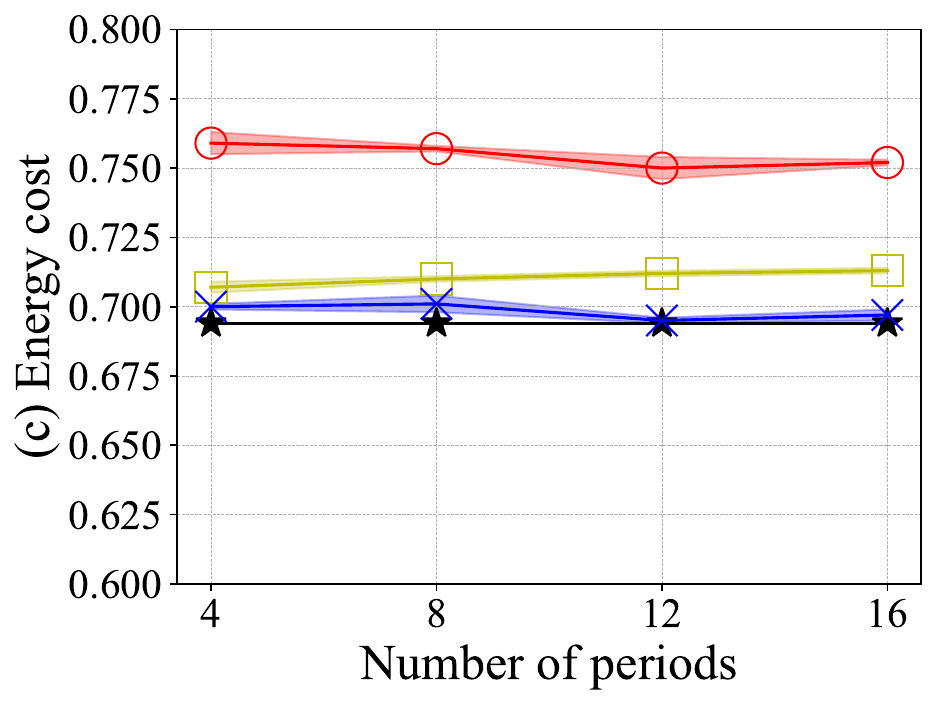}
		\label{fig:periods_energy}
	}
         \subfigure{
		\includegraphics[height=3.2cm,width=3.8cm]{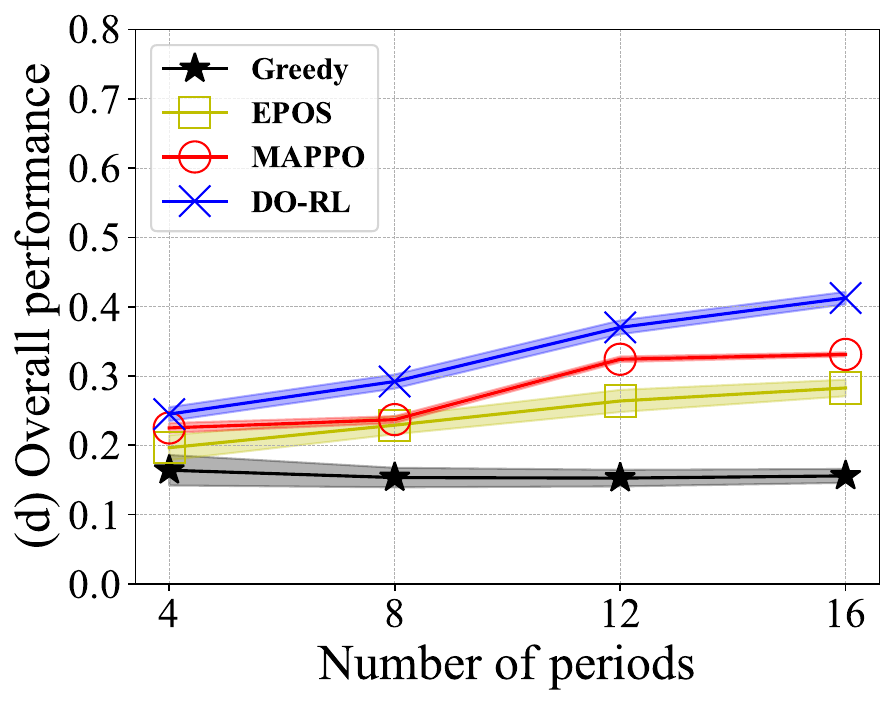}
		\label{fig:periods_overall}
	}
    \caption{\textbf{High number of periods increases both mission efficiency and sensing accuracy of all methods, especially for long-term learning methods (\emph{DO-RL} and \emph{MAPPO}).} DO-RL also shows its superior performance across different number of periods. Changing the the number of periods from $4$ to $16$ and fixing 16 drones, 64 cells, 4 charging stations and high density of vehicles.}
    \label{fig:periods}
\end{figure}

\begin{figure}[!t]
    \centering
    \includegraphics[width=\linewidth]{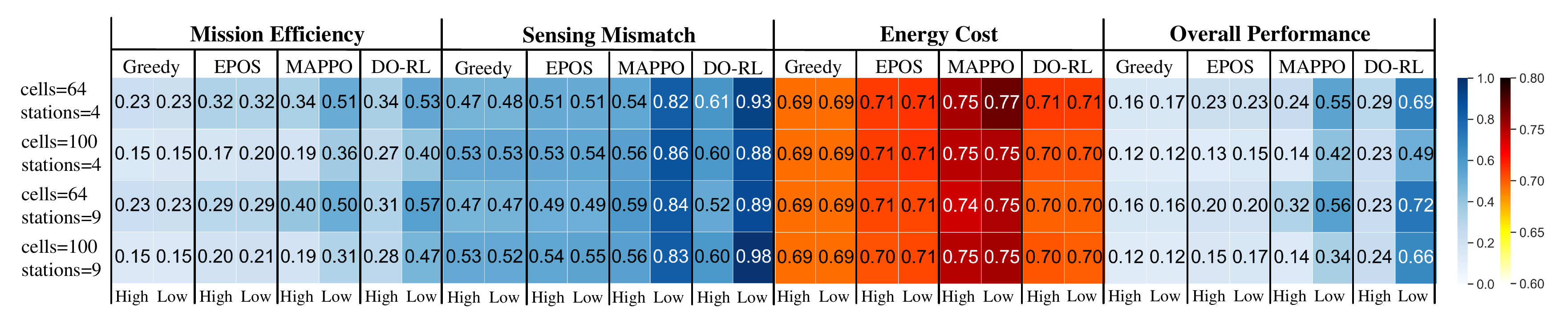}
    \caption{\textbf{Both low number of cells and high number of charging stations increase the overall performance of all methods. Long-term learning methods outperform short-term optimization methods in observing traffic flow when vehicles are sparsely distributed across the map.} Performance comparison under varying parameters: the number of cells ($64$ and $100$), the number of charging stations ($4$ and $9$), and the density of vehicles (high and low). The red palette on the right represents the values of energy cost, while the blue one denotes the values of mission efficiency, sensing mismatch and overall performance.}
    \label{fig:diff}
\end{figure}

\begin{figure}[!t]
	\centering
         \subfigure[High density of vehicles.]{
		\includegraphics[height=3.1cm,width=4.2cm]{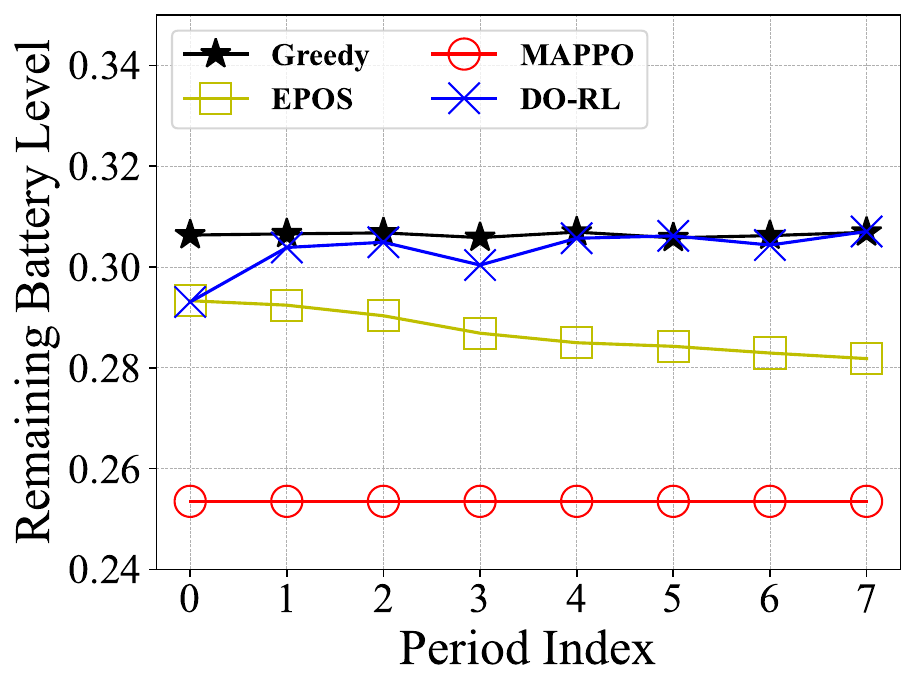}
		\label{fig:battery_u}
	}
         \subfigure[Low density of vehicles.]{
		\includegraphics[height=3.1cm,width=4.2cm]{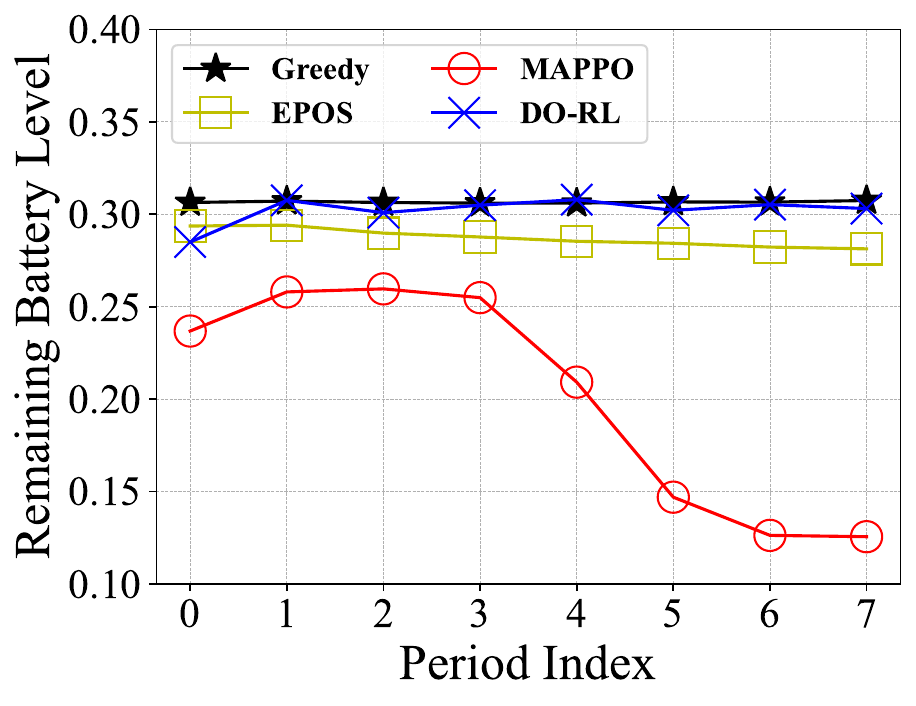}
		\label{fig:battery_s}
	}
    \caption{\textbf{\emph{DO-RL} keeps drones at a safe remaining battery level before recharging.} Performance comparison of the remaining battery level of four methods on both high and low density of vehicles.}
    \label{fig:battery}
\end{figure}

\begin{figure}[!t]
	\centering
        \subfigure[Drones Density=0.25, high density of vehicles.]{
		\includegraphics[height=3.1cm,width=3.8cm]{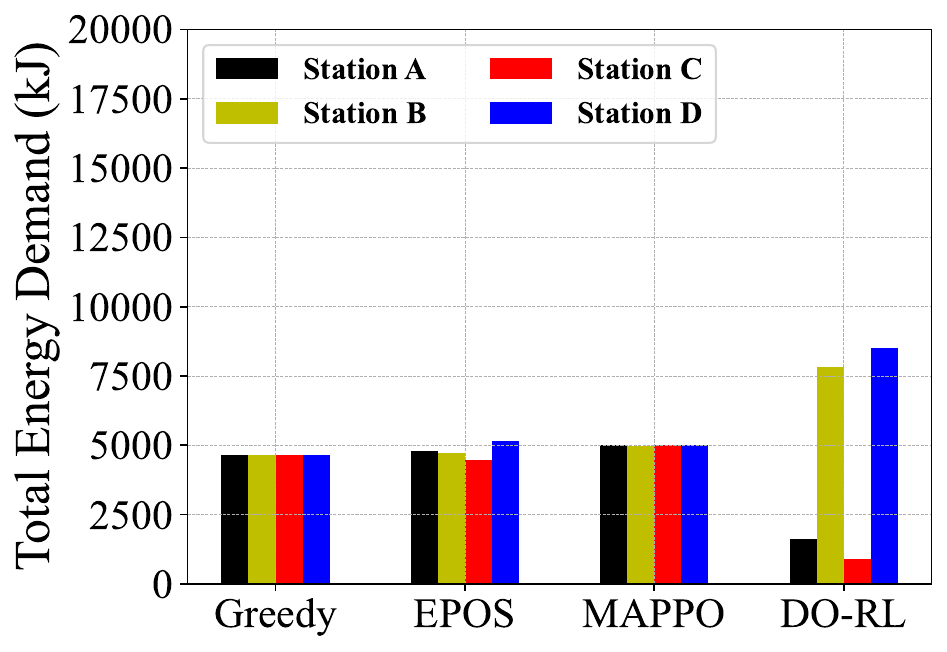}
		\label{fig:load_mu}
	}
         \subfigure[Drones Density=0.25, low density of vehicles.]{
		\includegraphics[height=3.1cm,width=3.8cm]{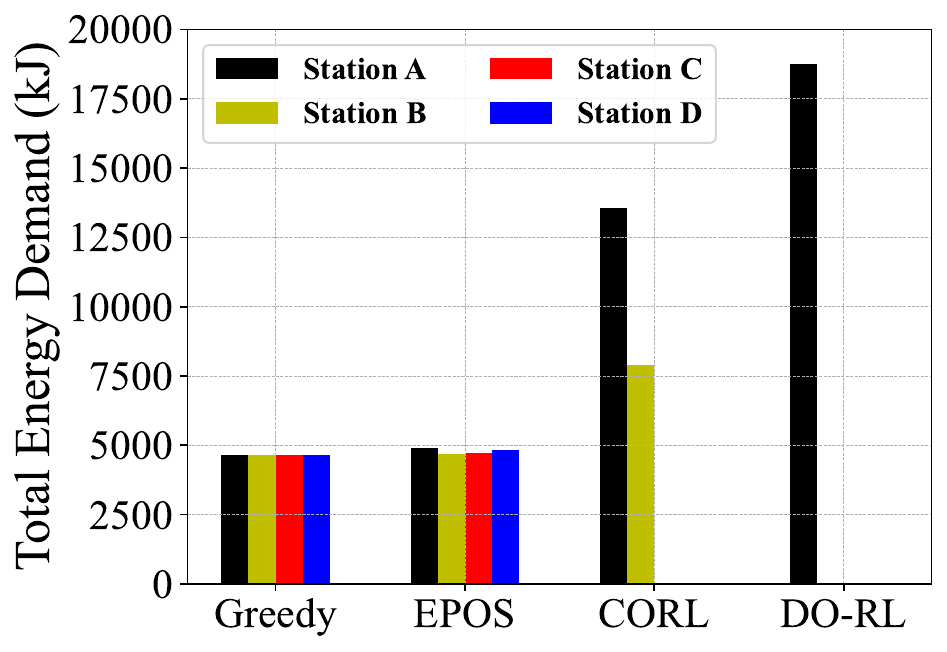}
		\label{fig:load_ms}
	}
         \subfigure[Drones Density for \emph{DO-RL}, high density of vehicles.]{
		\includegraphics[height=3.1cm,width=3.8cm]{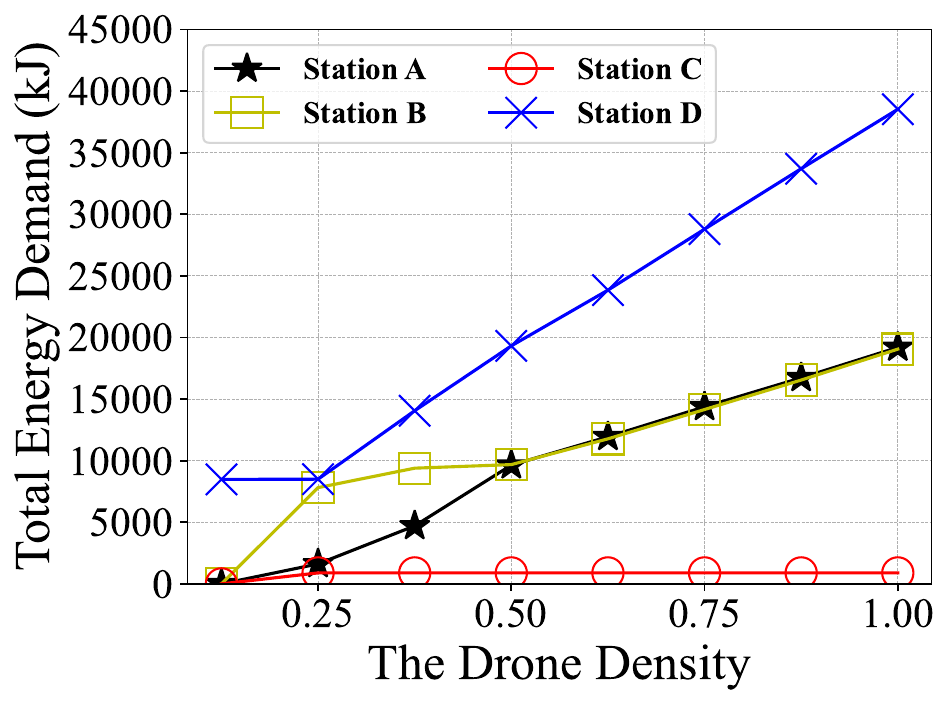}
		\label{fig:load_du}
	}
         \subfigure[Drones Density for \emph{DO-RL}, low density of vehicles.]{
		\includegraphics[height=3.1cm,width=3.8cm]{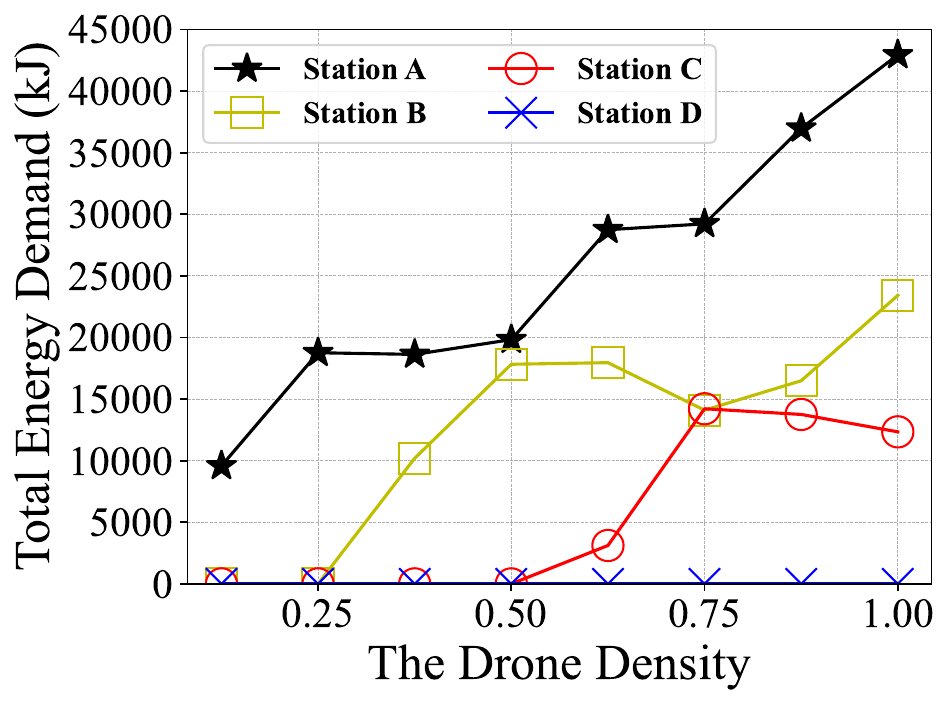}
		\label{fig:load_ds}
	}
    \caption{\textbf{\emph{DO-RL} relies on a lower number of charging places compared to other methods.} Performance comparisons of the total energy demand of four methods with $0.25$ density of drones, see (a) and (b), and the \emph{DO-RL} with different density of drones, see (c) and (d). Both comparisons are categorized by the high and low density of vehicles, see Fig.~\ref{fig:charging_dist}.}
    \label{fig:charge_load}
\end{figure}

To study the sensing scenario, there are six dimensions: (i) the density of drones, (ii) the number of periods, (iii) the number of cells, (iv) the number of charging stations, (v) the density of vehicles, (vi) the number of drones and cells while fixing the density of drones. Results are shown in Fig.~\ref{fig:drones}-\ref{fig:charge_load}, where the shadow area around the lines represents the standard deviation error of the results.

\cparagraph{Density of drones}
It denotes the ratio of the number of drones over the number of cells, representing the coverage of a single drone over the map. In this case, the number of cells is fixed as $64$, and increase the number of drones from $8$ to $64$, that is, the drones density increases from $0.125$ to $1.0$. Fig.~\ref{fig:drones} illustrates the performance of \emph{DO-RL} and baseline methods when using different drones density to perform traffic monitoring. As drones density increases, both mission efficiency and sensing accuracy of \emph{DO-RL} increase linearly. These are approximately $23.0\%$ and $15.8\%$ higher than \emph{EPOS} respectively, as shown in Fig.~\ref{fig:drones_effi} and Fig.~\ref{fig:drones_acc}~\footnote{\emph{DO-RL}, \emph{EPOS} and \emph{MAPPO} have similar overall performance when drones density is $1.0$, with maximum p-value of $0.04$, because there are ample drone resources available to effectively coordinate the coverage of the map.}. This is because \emph{DO-RL} controls the traveling direction of drones based on the predicted sensor data, so that drones cover the area with the maximum sensing value. \emph{MAPPO} has high efficiency and accuracy with a low density of drones (maximum p-value less than $0.001$ using Mann-Whitney U test), but its performance degrades when the drones density is higher than $0.375$ due to the high computational complexity. In contrast, the plan selection in \emph{DO-RL} reduces the action space and mitigates the learning difficulty, resulting in high performance with a high number of drones. In Fig.~\ref{fig:drones_energy}, the energy cost of \emph{DO-RL} is on average $1.7\%$ lower than \emph{EPOS} and $7.9\%$ lower than \emph{MAPPO}. \emph{Greedy} has the minimum energy cost among all methods, only $7.89\%$ lower than \emph{DO-RL}, but has the lowest performance in efficiency and accuracy. \emph{DO-RL} sacrifices a little energy for sensing, but still gets lower cost than other methods.

\cparagraph{Number of periods}
As shown in Fig.~\ref{fig:periods}, \emph{DO-RL} achieves superior performance compared to other methods, increasing linearly as the number of periods increase (increased by $51.3\%$ in quadruple time). Although it has statistically similar mission efficiency to \emph{MAPPO} when the number of periods is $4$, this metric increases dramatically because drones update and obtain more accurate predicted sensor data with higher number of periods. The timeslot-by-timeslot learning in \emph{MAPPO} perplexes the actions learned by drones compared to \emph{DO-RL}, since drones only take actions once per period ($30 min$) in \emph{DO-RL} but take actions per minute in \emph{MAPPO}. This also results in higher flying energy consumed by drones, and thus the energy cost of \emph{MAPPO} is around $8.1\%$ higher than \emph{DO-RL}.

Fig.~\ref{fig:diff} illustrates the performance comparison between \emph{DO-RL} and baseline methods, varying the number of cells, charging stations and the density of vehicles, while keeping the number of drones fixed at $16$ and periods at $8$.

\cparagraph{Number of cells}
If the number of cells increases from $64$ to $100$, the density of drones decreases. In Fig.~\ref{fig:diff}, \emph{DO-RL} shows a lower decrease in mission efficiency (decreased by $27.04\%$) than \emph{MAPPO} (decreased  by $44.50\%$) and \emph{EPOS} (decreased by $45.11\%$) as the number of cells increases. Furthermore, \emph{DO-RL} keeps relatively constant both sensing accuracy and energy cost as the number of cells varies.

\cparagraph{Number of charging stations}
If the number of charging stations increases from $4$ to $9$, the distance between the sensing areas and stations is reduced, cutting down the energy cost of recharging. As a result, the overall performance of both \emph{DO-RL} and \emph{EPOS} increase by $19.52\%$ and $14.35\%$ respectively. 

\cparagraph{Density of vehicles}
If a new map area that has low density of vehicles is selected, the distribution of sensor data changes (with different road distribution), as shown in Fig.~\ref{fig:charging_dist}. In Fig.~\ref{fig:diff}, both \emph{DO-RL} and \emph{MAPPO} have higher mission efficiency and sensing accuracy than the other two methods, with an average of $90.65\%$ higher efficiency and $76.85\%$ higher accuracy. This is because the learning methods observe the environment and control drones to collect data in cells and timeslots with the highest number of vehicles. \emph{DO-RL} performs better than other methods under low vehicle density even though the number of cells and charging stations increase.

Apart from sensing performance, the remaining battery level and charging load of \emph{DO-RL} in the sensing scenario is studied.

\cparagraph{Remaining battery level}
It denotes a percentage of battery level once drones have completed their sensing tasks. The value is calculated as an average among all drones within each time period. As shown in the Fig.~\ref{fig:battery_u} and \ref{fig:battery_s}, \emph{DO-RL} keeps drones at a high remaining battery level on high and low density of vehicles, with approximately $30.25\%$ and only $1.05\%$ lower than \emph{Greedy}. This level follows the drones' safety regulations which suggest finishing the missions when battery life is around $25\% - 30\%$. However, \emph{MAPPO} does not meet the regulations under low density of vehicles, and the minimum battery level can reach $12.55\%$. 

\cparagraph{Charging load}
It is the total energy demand of drones on each charging station over all time periods. This aims to study the placement of charging stations. As shown in the Fig.~\ref{fig:load_mu} and \ref{fig:load_ms}, the total energy demand on charging stations of \emph{DO-RL} over all periods is more imbalanced compared to other methods. This is because with \emph{DO-RL} drones learn to travel to the areas with the high required sensor data values. Since drones set the nearest charging stations as destinations, they rely on a lower number of charging places. For example, as shown in Fig.~\ref{fig:load_du} and \ref{fig:load_ds}, when the density of drones is $0.125$ under high density of vehicles, drones only depart from and return to the charging station $D$ since the vehicle distribution around $D$ is more dense than other charging stations. As the density of drones becomes $1$, drones gradually rely on other charging stations of $A$ and $B$. Similarly, drones only need to fly over the vehicle-dense areas around the charging station $A$ and $B$ under low density of vehicles. These results provide insights to policy makers for providing higher amount of energy on vehicle-dense areas to support drones' charging.

\begin{figure}[!t]
    \centering
    \includegraphics[width=\linewidth]{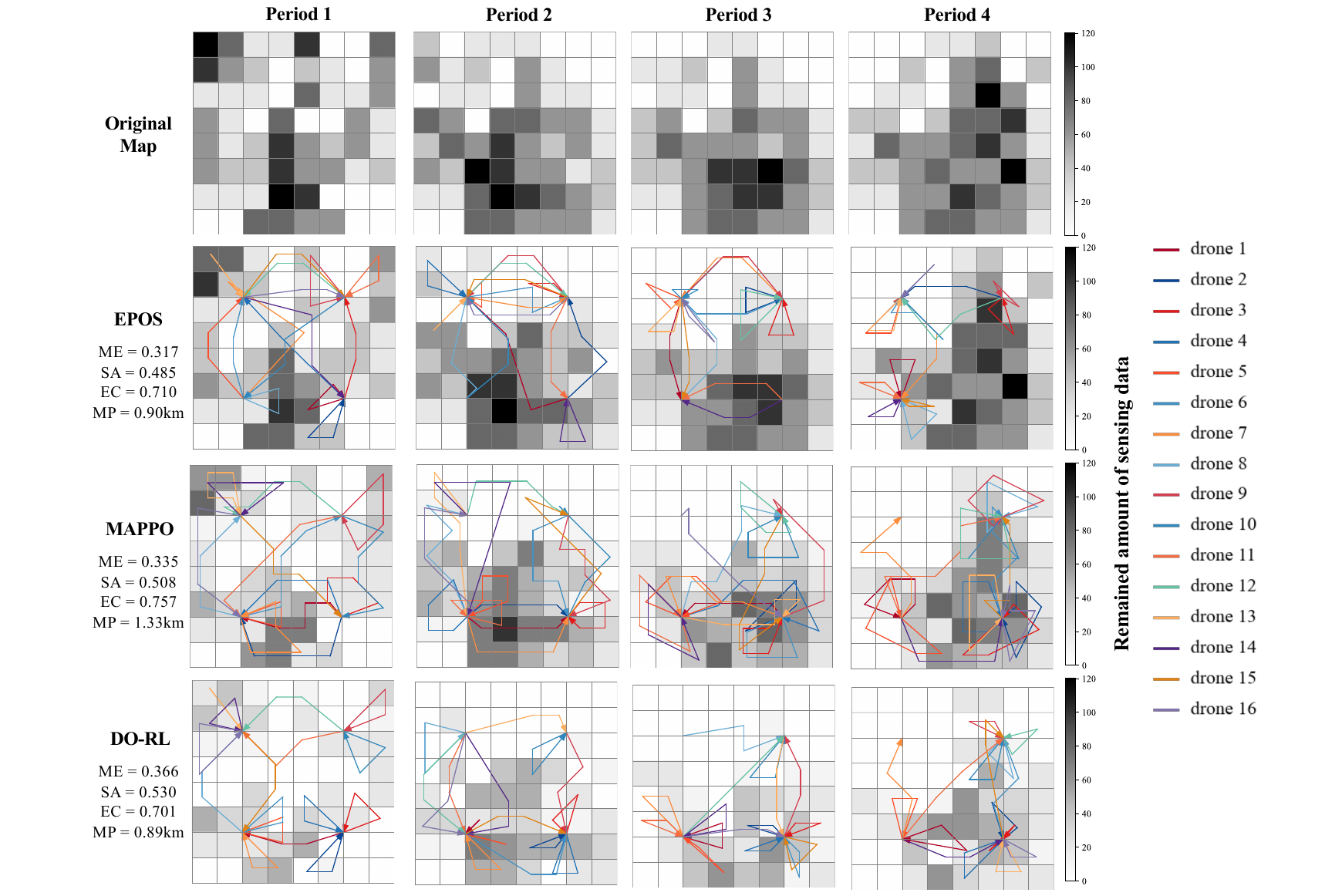}
    \caption{Comparison of optimal drone trajectories among all three approaches in different periods (ME is mission efficiency, SA is sensing accuracy, EC is energy cost, and MP is mean path length).}
    \label{fig:trajectories}
\end{figure}

Finally, we evaluate the drone trajectories of all methods over time. Fig.\ref{fig:trajectories} shows a visual comparison of optimal drone trajectories in \emph{DO-RL} compared to baseline methods. In this case of 16 drones fly to the grid cells and collect the required sensing data over first 4 of 8 time periods (each drone trajectory represents a color in Fig.\ref{fig:trajectories}). \emph{EPOS} fails to adapt to the changing sensing requirements and control drones to fly over the low-traffic areas, which results in higher remained amount of sensor data after period 3. As a consequence, \emph{EPOS} has lower mission efficiency and sensing accuracy than \emph{DO-RL} as illustrated in Section~\ref{sec:results}. Although \emph{MAPPO} effectively learns the dynamic sensing environment through deep neutral networks, it takes a high number of unnecessary actions to cover the cells with low sensing data, leading to higher path length and higher traveling energy consumption than \emph{DO-RL}. In contrast, \emph{DO-RL} overcomes the drawbacks of both \emph{EPOS} and \emph{MAPPO} by finding the high sensing areas energy-efficiently and accurately.

\begin{table}[!t]
	\centering
	\caption{Overall performance comparison of all approaches under different parameter settings.}  
	\label{table:comparison_all}
    \resizebox{\linewidth}{!}
    {
    \begin{tabular}{l|cc|cc|cc}  
		\toprule  
		\tabincell{l}{\textbf{Parameter settings} \\ \textbf{Approaches.:}}
        &\tabincell{l}{Drone density \\ Low (=0.25)} 
        &\tabincell{l}{Drone density \\ High (=1.0)} 
        &\tabincell{l}{Time periods \\ Low (=4)} 
        &\tabincell{l}{Time periods \\ High (=16)} 
        &\tabincell{l}{Vehicle density \\ High} 
        &\tabincell{l}{Vehicle density \\ Low} \\
		\midrule     
            \emph{Greedy}  &0.14 &0.62 &0.16 &0.16 &0.15 &0.16  \\
            \emph{EPOS}    &0.22 &1.78 &0.20 &0.28 &0.23 &0.23   \\
            \emph{MAPPO}   &0.23 &1.73 &0.23 &0.33 &0.24 &0.55  \\
            \emph{DO-RL}   &\textbf{0.28} &1.79 &0.25 &\textbf{0.41} &0.32 &\textbf{0.70}   \\
		\bottomrule
	\end{tabular}  
    }
\end{table} 

\subsection{Discussion and new insights}
In summary, the experimental results demonstrate the superior performance of the proposed approach \emph{DO-RL}. Several scientific insights on \emph{DO-RL} are listed as follows:

\cparagraph{\emph{DO-RL} achieves a win-win synthesis of short- and long-term strategies on multi-drone sensing performance in complex scenarios}
Table~\ref{table:comparison_all} shows that the hybrid approach has $27.83\%$ and $23.17\%$ higher overall performance than standalone \emph{EPOS} and \emph{MAPPO} respectively in the scenario of 16 drones, 8 time periods and high vehicle density. Under scarce drone resources (the low drone density), this approach is more energy-efficient and accurate by employing long-term learning methods for optimizing sensing. In contrast, under abundant drone resources (high drone density), short-term optimization methods alone suffice. Furthermore, the learning methods can direct drones towards regions with more vehicles to achieve highly efficient and accurate vehicle observation, especially in a low vehicle density area within a long time span. \emph{DO-RL} effectively combines the strengths of both short-term optimization and long-term learning methods to complement each other and improve adaptability in complex environments. 

\cparagraph{\emph{DO-RL} optimizes recharging assignment of drones to enhance the overall sensing performance}
By strategically learning flying directions, \emph{DO-RL} guides drones to recharge at stations located in high-traffic regions. Through continuous observation and prediction of sensing value distributions (i.e., traffic flow), \emph{DO-RL} dynamically selects both the departure and destination recharging stations for drones. This approach effectively constrains drone flight paths within areas of high sensing demand (see Fig.~\ref{fig:trajectories}). As a result, it achieves a $23.0\%$ improvement in mission efficiency and a $15.8\%$ increase in sensing accuracy compared to \emph{EPOS}, which lacks directional recharging strategies. Beyond performance gains, the optimized recharging assignment also offers valuable insights for urban charging infrastructure design and energy resource planning.

\cparagraph{\emph{DO-RL} improves the drone-based sustainable sensing by reducing the energy consumption} 
The collective learning in \emph{DO-RL} mitigates the inefficiencies of conventional DRL by discouraging drones from consuming energy on sensing low-traffic regions. This not only lowers the energy cost of aerial sensing (around $7.9\%$ lower than \emph{MAPPO}), but also helps drones maintain battery levels above safety thresholds. As a result, this approach supports longer, safer and more sustainable missions. In the context of traffic monitoring, drones detect congestion early and traffic operators intervene to alleviate it. By contributing to more effective congestion management, \emph{DO-RL} contributes to carbon emission reduction, aligning with meeting net zero targets~\cite{bukhari2023zero}.

\section{Conclusion and Future Work}
This paper studies a autonomous task allocation problem for large-scale, dynamic spatio-temporal sensing by a swarm of autonomous drones. A new hybrid optimization approach named \emph{DO-RL} is proposed, by integrating both distributed optimization and DRL, to solve the problem by optimizing the flight paths, data collection and recharging assignment of drone swarms. In each time period, drones determine their overall flying directions to destinations with charging stations via DRL, and autonomously evolve their navigation and sensing plans using collective learning. Extensive experimentation using realistic urban mobility reveals the effectiveness of \emph{DO-RL} and provides valuable insights for the broader community: The combination of short-term and long-term strategies proves highly effective in overcoming their standalone limitations in drone-based traffic monitoring. Short-term methods compensate the challenges of training complexity, energy consumption and recharging, while long-term approaches account for maximizing accumulated rewards, enhancing the overall sensing performance. The proposed approach allows the system to dynamically adapt to changing environments, efficiently managing varying drone resources and enabling extended operations.

However, this hybrid approach comes with limitations and can be further improved towards several research avenues: (i) It relies on centralized training, which introduces potential system vulnerabilities to adversarial attacks. To mitigate this, future work should explore fully decentralized training paradigms~\cite{omidshafiei2017deep}, where agents learn and update strategies without centralized control. (ii) Decentralized coordination still faces security and trust challenges during inter-agent communication. Ensuring that drones share only essential and trustworthy information without exposing sensitive data is critical. Emerging techniques such as blockchain-based consensus, federated learning, or zero-knowledge proofs could help establish secure and verifiable cooperation among drones, promoting both robustness and privacy~\cite{wang2020optimizing,hao2025uav}. (iii) The practical constraints on multi-drone task allocation requires further exploration, such as obstacles (e.g., buildings), wind speed, or no-fly zones. Although previous work~\cite{qin2024m} introduces collision avoidance using artificial potential fields, this operates after the task allocation stage and may lead to plan violations. To address this, future research should focus on integrating environmental constraints directly into the task planning and selection process. (iv) The proposed approach should combine multiple types of sensors (e.g., cameras, LiDAR, thermal) to enable multi-modal sensing for more comprehensive urban data collection by heterogeneous drones, rather than tailored to traffic monitoring application.

\section*{ACKNOWLEDGEMENTS}
This research is supported by a UKRI Future Leaders Fellowship (MR-/W009560/1): 
\emph{Digitally Assisted Collective Governance of Smart City Commons–ARTIO}, and the European Union, under the Grant Agreement GA101081953 for the project H2OforAll — \emph{Innovative Integrated Tools and Technologies to Protect and Treat Drinking Water from Disinfection Byproducts (DBPs)}. Views and opinions expressed are, however, those of the author(s) only and do not necessarily reflect those of the European Union. Neither the European Union nor the granting authority can be held responsible for them. Funding for the work carried out by UK beneficiaries has been provided by UKRI under the UK government’s Horizon Europe funding guarantee [grant number 10043071]. Thanks to Manos Chaniotakis and Zeinab Nezami for the support on the transportation dataset.

\bibliographystyle{unsrt}

\bibliography{reference}

\appendix
\section{Effect of different parameters} \label{appendix_param}
The objective of this appendix is to analyze the impact of various parameters (defined in Section~\ref{sec:ex_settings}) within the plan generation, distributed optimization and reinforcement learning of the proposed method. The insights derived from this appendix can be used to make informed empirical decisions regarding parameter selection for the evaluation scenarios. The calculations presented herein can be automated for any scenario during hyper-parameter optimization. Such optimization is not the focus of this paper.

We use the \emph{DO-RL} and the basic scenario to test the mobility range, the agents' behavior, optimization objective function, and hyperparameter. 

\begin{figure}[htbp]
	\centering
        \subfigure[Mobility range.]{
		\includegraphics[height=3.6cm,width=4.5cm]{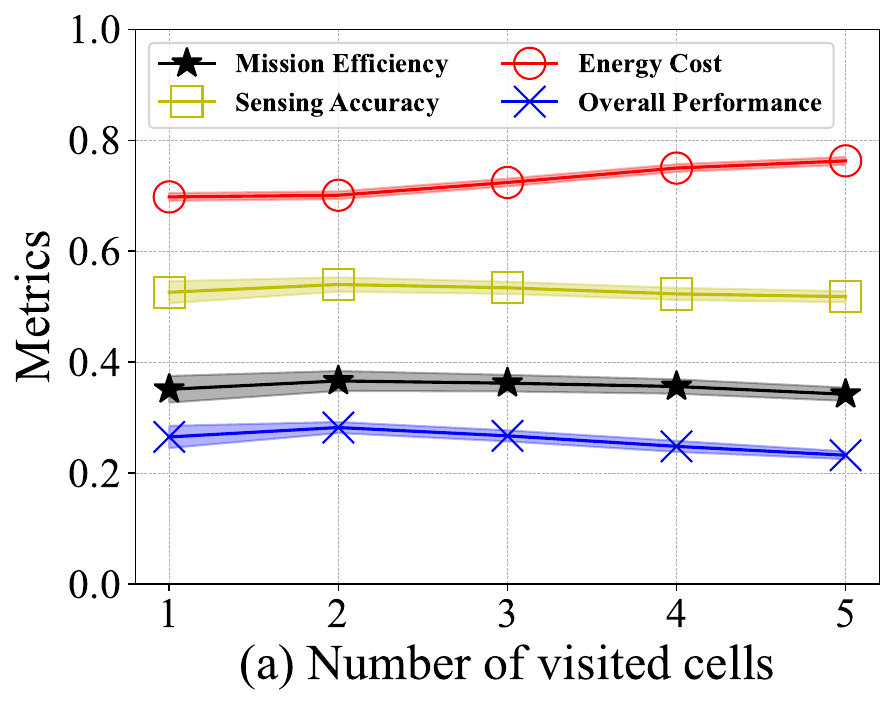}
		\label{fig:mobility}
	}
         \subfigure[Agents' behavior.]{
		\includegraphics[height=3.6cm,width=4.5cm]{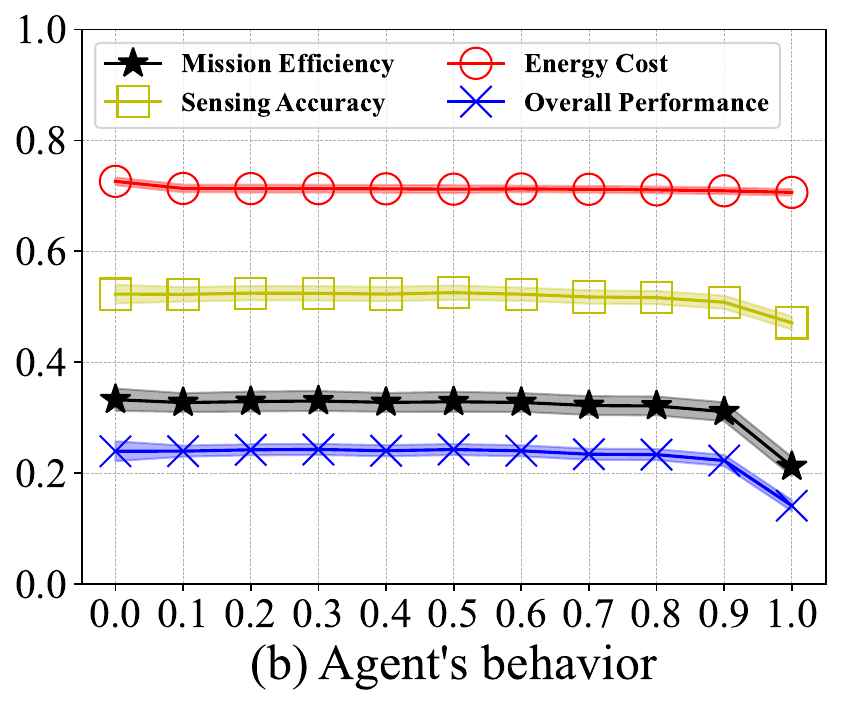}
		\label{fig:behavior}
	}
        \subfigure[Optimization objective function.]{
		\includegraphics[height=3.6cm,width=5cm]{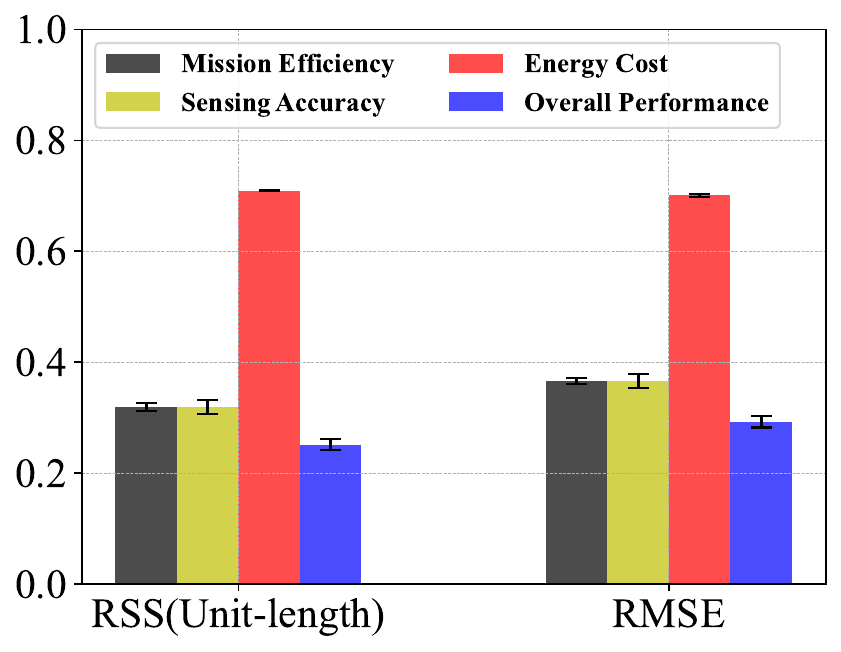}
		\label{fig:objective}
	}
    \caption{\textbf{\emph{DO-RL} has a superior overall performance when $J_u = 2$, $ \beta \in [0.1, 0.8] $, and using RMSE.} Change the parameters of the proposed method in mobility range, agents' behavior and optimization objective function.}
    \label{fig:param}
\end{figure}

\cparagraph{Mobility range}
Fig.~\ref{fig:mobility} illustrates the effect of different numbers of visited cells. The higher the number of cells a drone visits, the higher the flying energy is, and the higher the energy cost is. The high number of visited cells perplexes the spatio-temporal navigation and sensing of a drone, leading to over-sensing. Specifically, multiple drones visit the same cell simultaneously and waste energy on collecting the same data. In contrast, if drones visit only one cell, they are free from over-sensing, but have a high probability to miss the area with a high required sensing data due to its low mobility range. As a result, we empirically select $J(a^u) = 2$ for each drone to optimize the overall performance. 

\cparagraph{Agents' behavior}
Fig.~\ref{fig:behavior} shows the effect of agents' behavior by varying the parameter of $\beta$~\cite{pournaras2018decentralized}. As $\beta$ increases from $0$ to $1$, agents reduce the energy cost of their selected plans, while decreasing both efficiency and accuracy. This is because, according to Eq.~\ref{eq:epos}, drones with higher $\beta$ choose a plan with lower energy cost, which degrades the matching between the total sensing and the target. Since the overall performance for $\beta \in [0.1, 0.8]$ is statistically similar, using the Mann-Whitney U test with average p-value of $0.0003$, drones are allowed to choose their behaviors within this range.

\cparagraph{Optimization objective function}
Fig.~\ref{fig:objective} illustrates the comparison of two different cost functions used in EPOS. Apart from RMSE, we choose the correlation error, i.e., the residual sum of squares (RSS) between the aggregated plans and the target, both in unit-length scaled. The results show that RSS has lower efficiency and accuracy, and thus has lower overall performance than RMSE. The low p-value (less than $0.001$) proves that the difference is statistically significant.

\cparagraph{Hyperparameter in deep reinforcement learning}
Table~\ref{table:training} shows the training performance of different hyperparameter in DRL, including the discount factor $\gamma$, batch size $H$, clip interval $\epsilon$, and deep neutral networks. It compares the reward and the episode to converge by changing these parameters. The results demonstrate a more stable (lower reward error and higher converged episode) but lower average reward in training when decreasing $\gamma$, $\epsilon$ or increasing $H$. In contrast, a relatively higher average reward leads to higher error and slower convergence. In addition, the RNN used in the proposed \emph{DO-RL} performs better than the multilayer perceptron (MLP), with $7.8\%$ higher average reward. Considering these results, the parameters of $\gamma = 0.95$, $H = 64$, $\epsilon=0.2$, and RNN neutral network are selected for \emph{DO-RL}.

\begin{table}[htbp]
	\centering
	\caption{Reward and convergence vs. hyperparameter. The proposed approach takes the parameters of $\gamma = 0.95$, $H = 64$, $\epsilon=0.2$, and RNN neutral network.}  
	\label{table:training}
    \resizebox{14cm}{!}
    {
    \begin{tabular}{lcccccccc}  
		\toprule  
		\tabincell{l}{\textbf{Approaches} \\ \textbf{Attributes.:}}
        &\tabincell{l}{Proposed} 
        &\tabincell{l}{$\gamma = 0.90$} 
        &\tabincell{l}{$\gamma = 0.99$} 
        &\tabincell{l}{$H = 32$} 
        &\tabincell{l}{$H = 128$}
        &\tabincell{l}{$\epsilon = 0.1$}
        &\tabincell{l}{$\epsilon = 0.3$}
        &\tabincell{l}{MLP} \\
		\midrule     
        \emph{Average reward}       &8.42 &8.45 &8.24 &8.30 &8.42 &8.17 &8.34 &7.81  \\
        \emph{Error of reward}         &0.68 &0.79 &0.66 &0.66 &0.71 &0.68 &0.87 &0.55  \\
        \emph{Converged episode}    &1899 &2048 &2451 &1071 &2589 &1843 &2116 &1657   \\
		\bottomrule
	\end{tabular}  
    }
\end{table}

\section{Additional comparison of approaches} \label{appendix_compare}
This appendix extends the comparison results between the proposed \emph{DO-RL} and baseline methods. The performance evaluation contains the following aspects: (i) computational and communication overheads; (ii) training convergence, and (iii) the number of drones and cells while fixing the density of drone.

\begin{figure}[htbp]
    \centering
        \subfigure[Computational cost vs. drone density.]{
		\includegraphics[height=3.2cm,width=3.8cm]{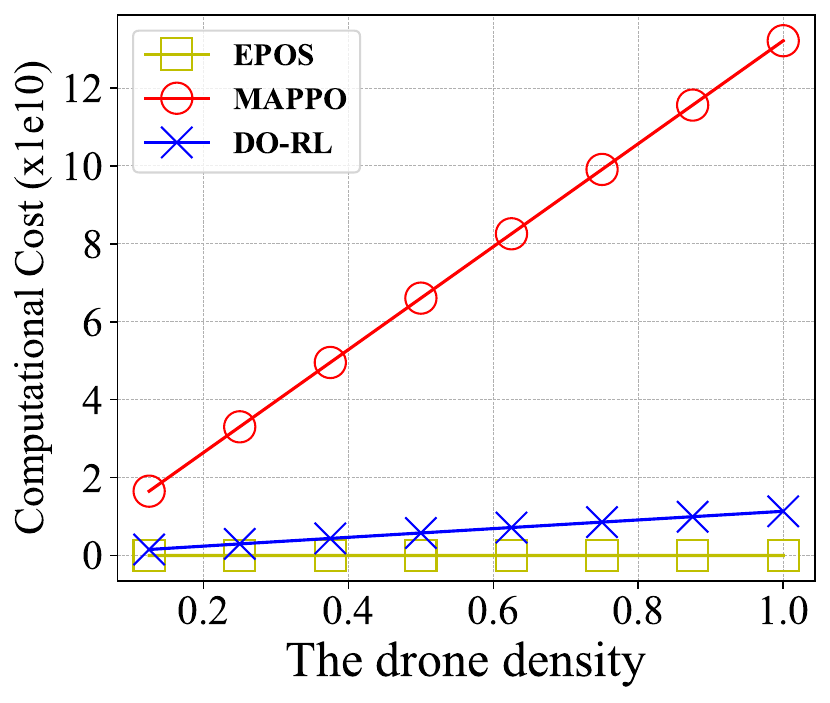}
		\label{fig:computation_agents}
	}
        \subfigure[Computational cost vs. number of periods.]{
		\includegraphics[height=3.2cm,width=3.8cm]{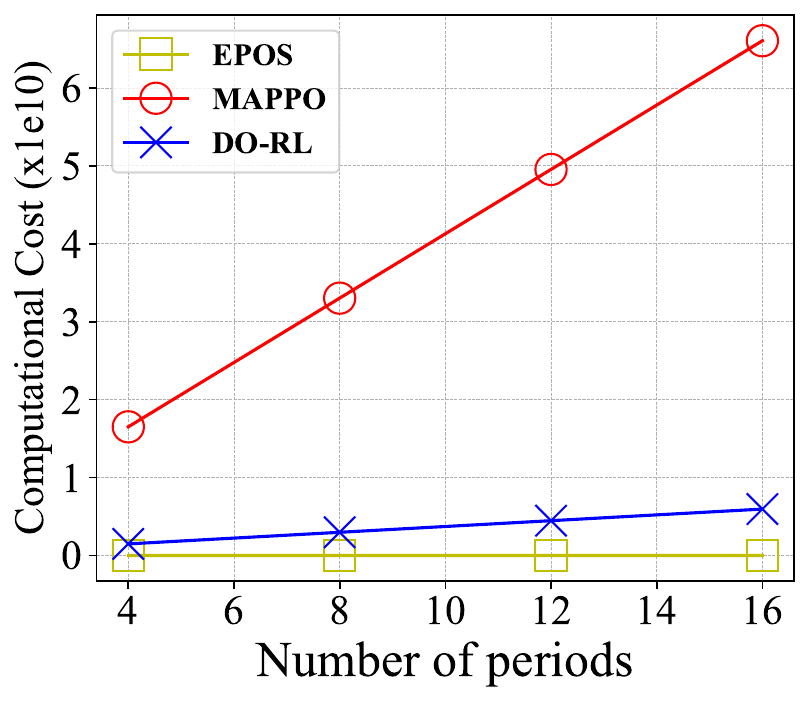}
		\label{fig:computation_periods}
	}
        \subfigure[Communication cost of \emph{DO-RL} vs. drone density.]{
		\includegraphics[height=3.2cm,width=3.8cm]{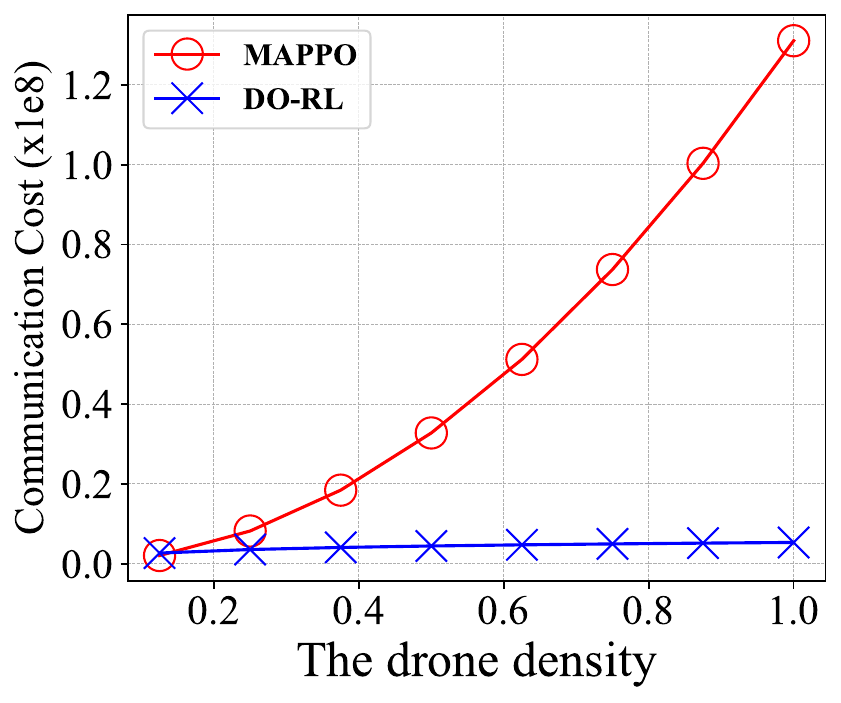}
		\label{fig:communication_proposed}
	}
        \subfigure[Communication cost of \emph{EPOS} vs. drone density.]{
		\includegraphics[height=3.2cm,width=3.8cm]{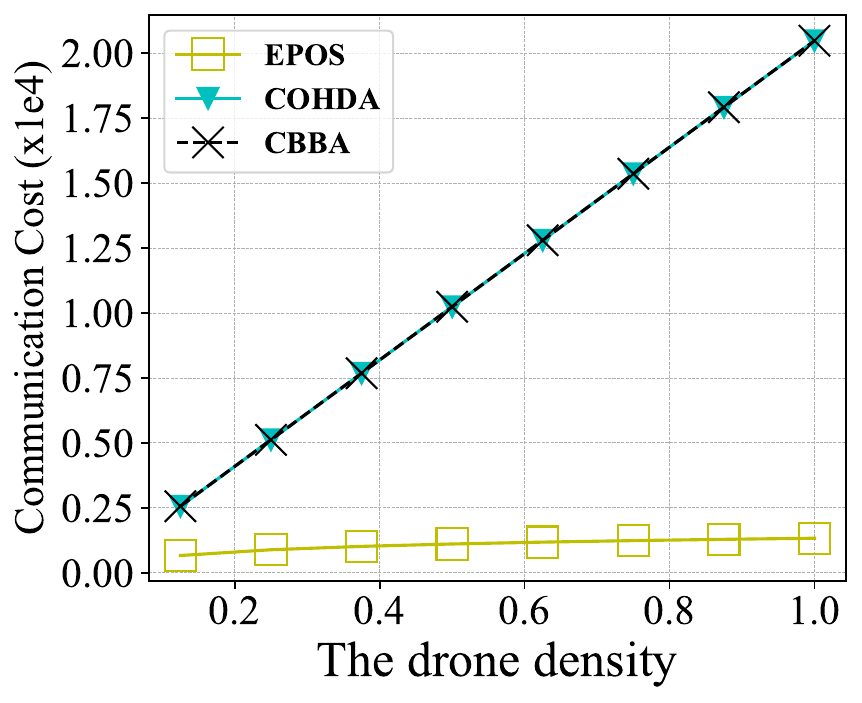}
		\label{fig:communication_epos}
	}
    \caption{Comparison of computational and communication overheads of all methods.}
    \label{fig:complexity}
\end{figure}

\cparagraph{Computational and communication overheads}
Fig.\ref{fig:complexity} illustrates that \emph{DO-RL} significantly reduces computational and communication cost during training compared to \emph{MAPPO}. As the drone density and number of periods increase, \emph{MAPPO} exhibits a near-linear growth in computation. The tree communication structure in \emph{DO-RL} also decreases the communication cost of drones by around $93.6\%$ when the drone density is $1$. Furthermore, Fig.\ref{fig:communication_epos} proves the lower communication overhead of the collective learning (EPOS) used in \emph{DO-RL} compared to other distributed optimization methods such as COHDA and CBBA.

\begin{figure}[htbp]
	\centering
        \subfigure[Reward change of \emph{DO-RL} with training episodes.]{
		\includegraphics[width=0.45\linewidth]{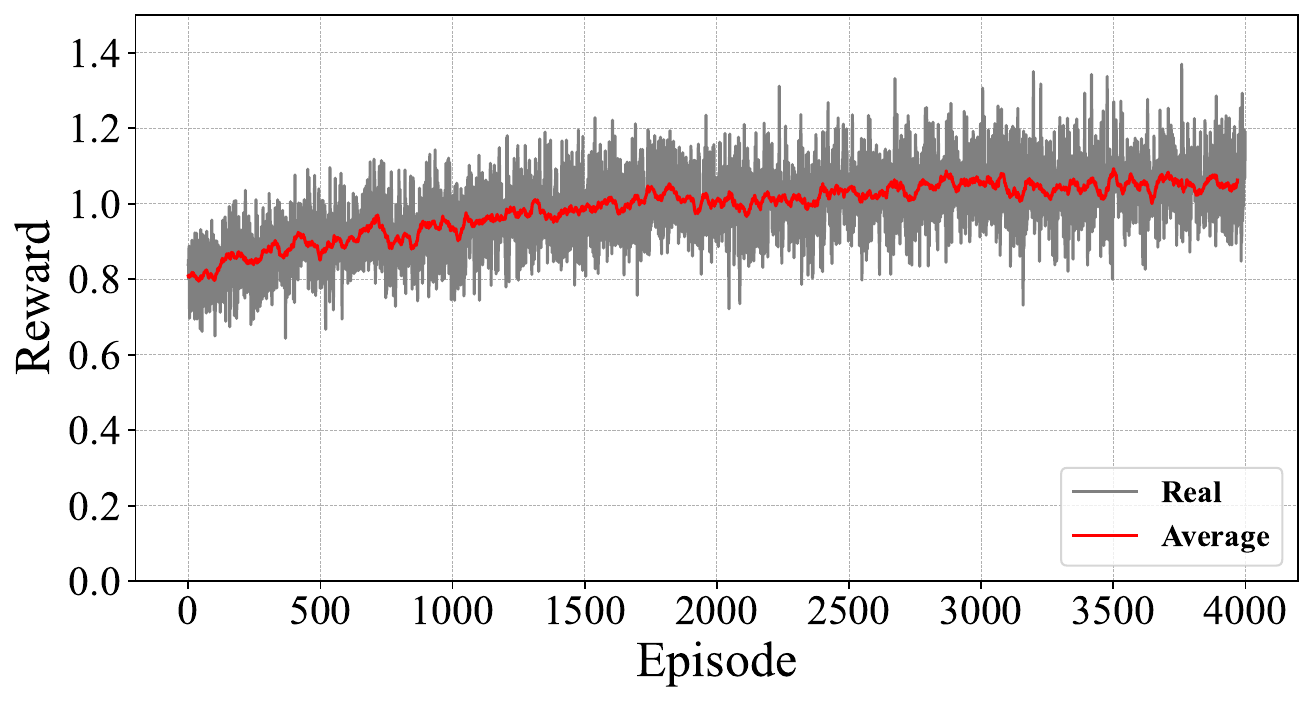}
		\label{fig:train_proposed}
	}
         \subfigure[Reward change of \emph{MAPPO} with training episodes.]{
		\includegraphics[width=0.45\linewidth]{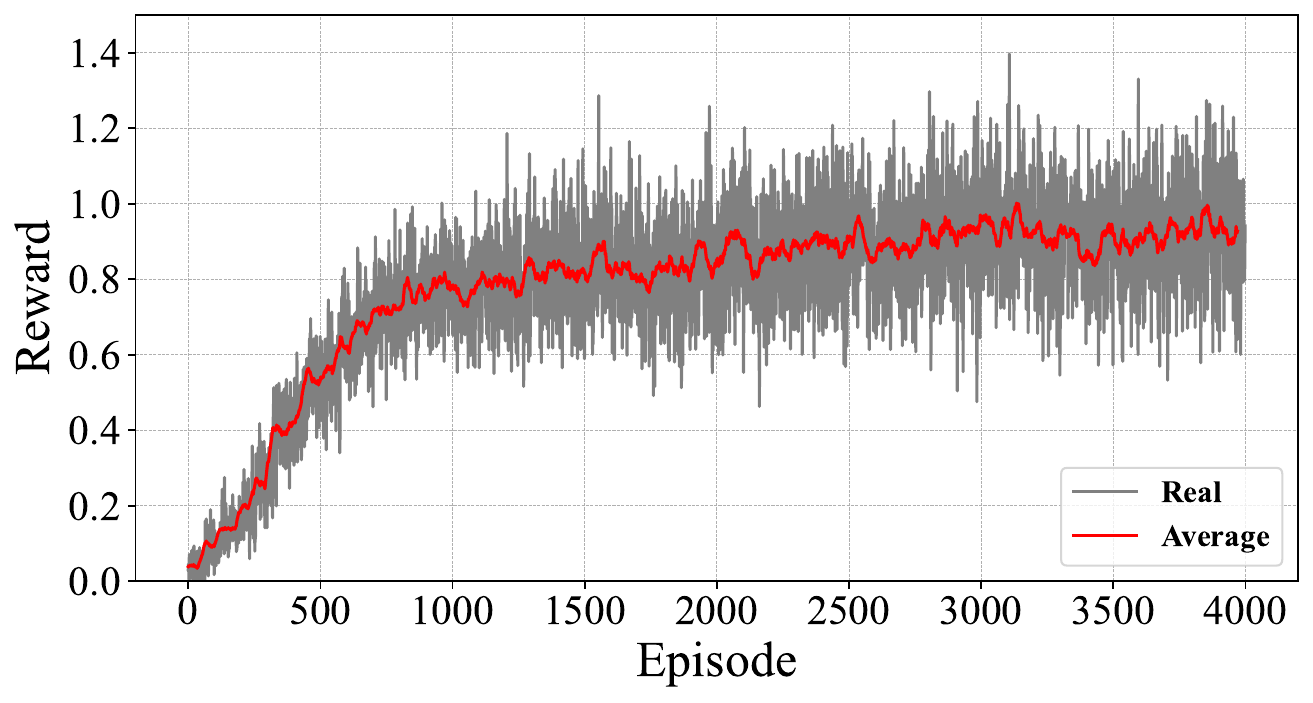}
		\label{fig:train_mappo}
	}
    \caption{The training process of \emph{DO-RL} and \emph{MAPPO}. The gray is the real value and red is the average value.}
    \label{fig:training}
\end{figure}

\cparagraph{Training convergence}
Fig.\ref{fig:training} compares the training of both \emph{DO-RL} and \emph{MAPPO}. Using collective learning, \emph{DO-RL} achieves faster exploration and converges in around $1899$ episode. It is significantly earlier than \emph{MAPPO}, which converges after $3000$ episode due to high training complexity. This further helps \emph{DO-RL} to explore a higher average reward than \emph{MAPPO}.

\begin{figure}[htbp]
	\centering
        \subfigure{
		\includegraphics[height=3.2cm,width=3.8cm]{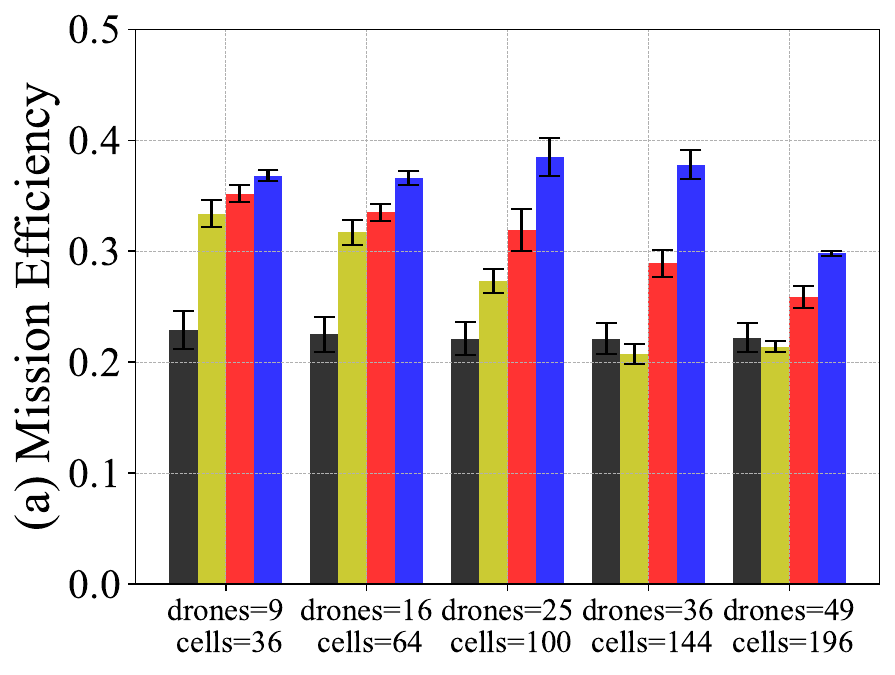}
		\label{fig:fixDensity_effi}
	}
         \subfigure{
		\includegraphics[height=3.2cm,width=3.8cm]{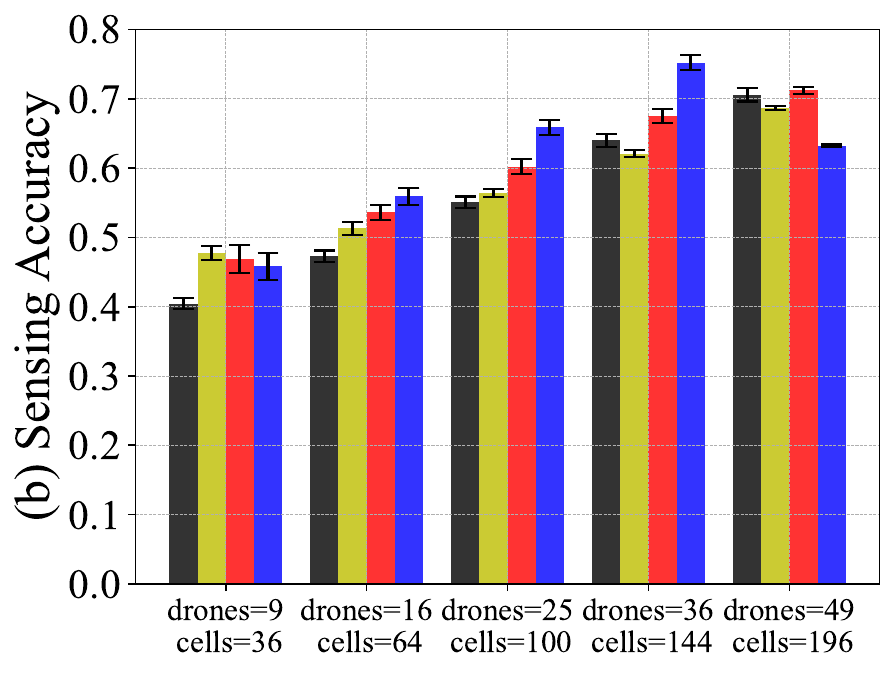}
		\label{fig:fixDensity_acc}
	}
         \subfigure{
		\includegraphics[height=3.2cm,width=3.8cm]{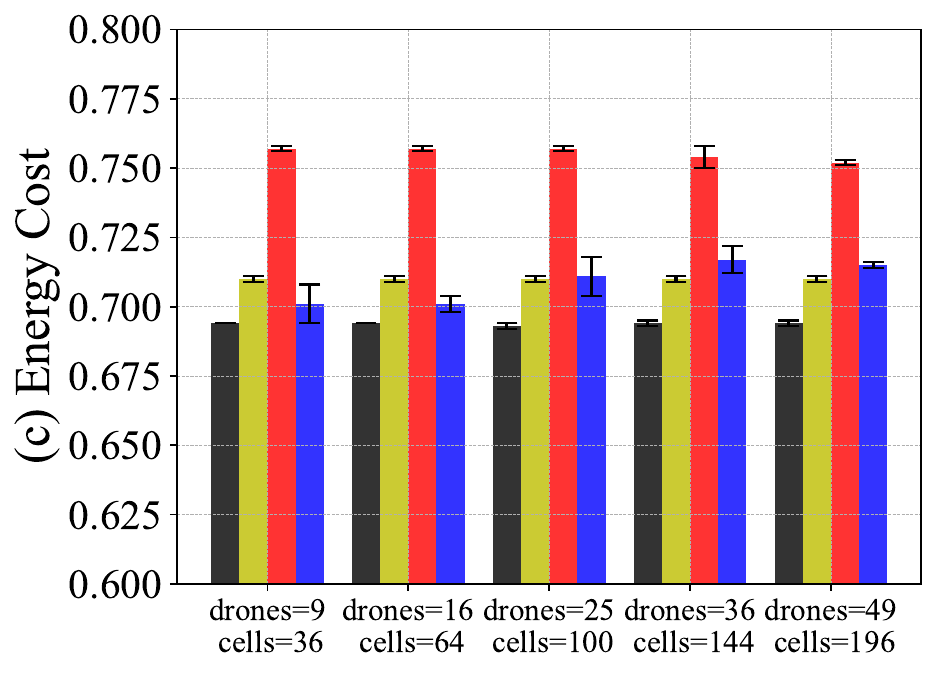}
		\label{fig:fixDensity_energy}
	}
         \subfigure{
		\includegraphics[height=3.2cm,width=3.8cm]{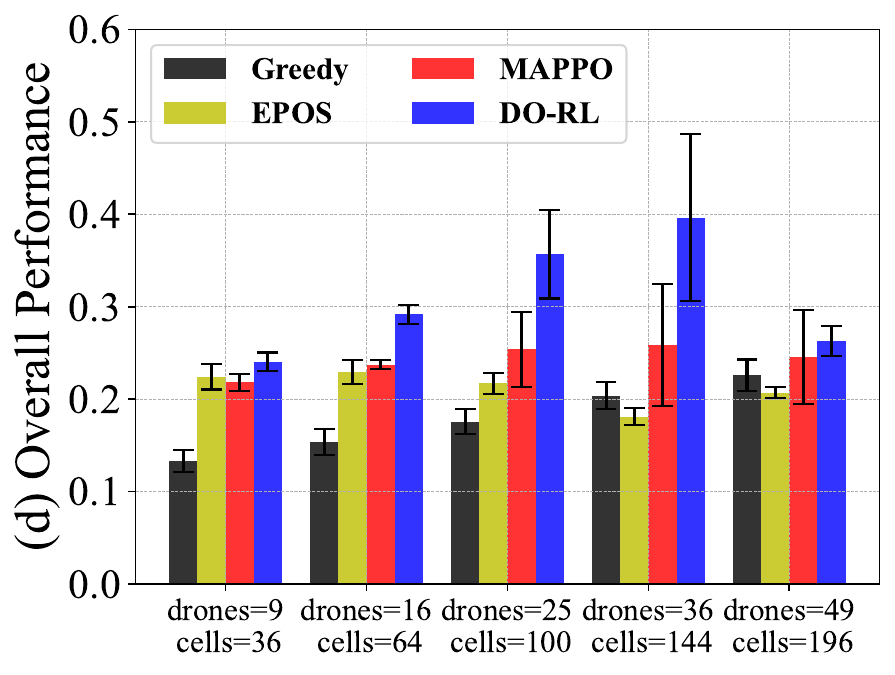}
		\label{fig:fixDensity_overall}
	}
    \caption{\textbf{High number of drones/cells with fixed drones density increases the sensing accuracy of all methods, peaking at $36$ drones and $144$ cells.} Changing the number of drones and cells but fixing the drones density at $0.25$ with 8 periods, 4 charging stations and high density of vehicles.}
    \label{fig:fixDensity}
\end{figure}

\cparagraph{Number of drones/cells with fixed drones density}
If we fix the density of drones but increase the number of drones, the number of cells increases proportionally. For example, when the density of drones is $0.25$ and the number of drones increases from $9$ to $49$, and then the number of cells should increase from $36$ to $196$. As shown in Fig.~\ref{fig:fixDensity}, when the number of drones/cells increases, the sensing accuracy of all methods increases. However, the mission efficiency of both \emph{EPOS} and \emph{MAPPO} decreases (by $35.93\%$ and $26.42\%$ respectively). This is due to \emph{EPOS} facing challenges in efficiently scheduling drones for a large number of cells, while \emph{MAPPO} experiences increased computational complexity. In contrast, \emph{DO-RL} overcomes these challenges and demonstrates a $4.62\%$ increase in mission efficiency. When increasing to $49$ drones and $196$ cells, however, the sensing accuracy of \emph{DO-RL} degrades, decreased by $15.96\%$, due to the high complexity of scheduling and computation. In overall, \emph{DO-RL} has statistically higher overall performance than other methods, with an average improvement of $46.98\%$ (maximum p-value less than $0.001$), which proves the efficiency of the proposed approach.

\section{Power consumption model}
Drones spend energy to surpass gravity force and counter drag forces due to wind and forward motions. A drone controls the speed of each rotor to achieve the thrust $T$ and pitch $\theta$ necessary to stay aloft and travel forward at the desired velocity while balancing the weight and drag forces. For a drone with mass $m_{b}$ and its battery with mass $m_{c}$, we define the total required thrust as follows:
\begin{equation}
	\mathcal{T} = (m_{b} + m_{c}) \cdot g + F_{d},
\end{equation}
\noindent where $g$ is the gravitational constant, and $F_{d}$ is the drag force that depends on air speed and air density. For steady flying, the drag force can be calculated by the pitch angle $\theta$ as:
\begin{equation}
	F_{d} = (m_{b} + m_{c}) \cdot g \cdot tan(\theta).
\end{equation}

Based on the model in~\cite{monwar2018optimized}, the power consumption with forward velocity and forward pitch is given by:
\begin{equation}
	C^\mathsf{f} = (v \cdot sin\theta + v_{i}) \cdot \frac{\mathcal{T}}{\epsilon},
	\label{power_flight}
\end{equation}
\noindent where $v$ is the average ground speed; $\epsilon$ is the overall power efficiency of the drone;  $v_{i}$ is the induced velocity required for given $T$ and can be found by solving the nonlinear equation:
\begin{equation}
	v_{i} = \frac{2 \cdot \mathcal{T}}{\pi \cdot d^{2} \cdot r \cdot \rho \cdot \sqrt{(v \cdot cos\theta)^{2} + (v \cdot sin\theta + v_{i})^{2}}},
\end{equation} 
\noindent where $d$ and $r$ are the diameter and number of drone rotors; $\rho$ is the density of air. 

Moreover, the power consumption for hovering of a drone is calculated by:
\begin{equation}
	C^\mathsf{h} = \frac{\mathcal{T}^{3/2}}{\epsilon \cdot \sqrt{\frac{1}{2} \pi \cdot d^{2} \cdot n \cdot \rho}}.
	\label{power_hover}
\end{equation}


\end{document}